\documentclass[lettersize,journal]{IEEEtran}
\usepackage{stfloats}
\usepackage{cite}
\usepackage{float}
\usepackage{bm}
\usepackage{here}
\usepackage{soul, color}
\usepackage{multirow}
\usepackage{tabularx}
\usepackage{algorithmic}
\usepackage{makecell}
\usepackage{hhline}
\usepackage{nicefrac}
\usepackage{lipsum}
\usepackage{array}
\usepackage{booktabs}
\usepackage{paralist}
\usepackage{gensymb}
\usepackage{inputenc}
\usepackage[dvipsnames]{xcolor}
\usepackage{colortbl}
\usepackage{svg}
\usepackage{threeparttable}
\usepackage{enumerate}
\usepackage{upgreek}
\usepackage[T1]{fontenc}
\usepackage{babel}
\usepackage{array}
\usepackage{textcomp}
\usepackage{mathtools}
\usepackage{amsmath}
\usepackage{amssymb}
\usepackage{graphicx}
\usepackage[unicode=true,
 bookmarks=true,bookmarksnumbered=true,bookmarksopen=true,bookmarksopenlevel=1,
 breaklinks=false,pdfborder={0 0 0},pdfborderstyle={},backref=false,colorlinks=false]
 {hyperref}
 
\makeatletter
\AtBeginDocument{
 \hypersetup{
 pdftitle = {\@title},
 pdfauthor = {\@author}
 }
}
\makeatother

\usepackage[caption=false,font=normalsize,labelfont=sf,textfont=sf]{subfig}
\usepackage{enumitem}

\title{Piezoelectric Soft Robot Inchworm Motion by Tuning Ground Friction through Robot Shape: Quasi-Static Modeling and Experimental Validation}
\author{Zhiwu Zheng, Prakhar Kumar, Yenan Chen, Hsin Cheng, \\Sigurd Wagner,
Minjie Chen, Naveen Verma and James C. Sturm\thanks{The authors thank Prof. M. Haataja of Princeton University for his advice on mechanical modeling. This work was supported by the Semiconductor Research Corporation
(SRC), DARPA, Princeton Program in Plasma Science and Technology,
and Princeton University. \emph{(Corresponding author: James C. Sturm)}}\thanks{The authors are with the Department of Electrical and Computer Engineering,
Princeton University, Princeton, New Jersey 08544, U.S.A. (e-mail:
{zhwzheng.1994@gmail.com}; prakhark@princeton.edu; yenanc@zju.edu.cn; hsin@princeton.edu; wagner@princeton.edu; minjie@princeton.edu; nverma@princeton.edu;
sturm@princeton.edu).}}

\begin{document}
\maketitle
\begin{abstract}

Electrically-driven soft robots based on piezoelectric actuators may enable compact form factors and maneuverability in complex environments. In most prior work, piezoelectric actuators are used to control a single degree of freedom. In this work, the coordinated activation of five independent piezoelectric actuators, attached to a common metal foil, is used to implement
inchworm-inspired crawling motion in a robot that is less than 0.5 mm thick. The motion is based on the control of its friction to the ground through the robot's shape, in which one end of the robot (depending on its shape) is anchored to the ground by static friction, while the rest of its body expands or contracts. A complete analytical model of the robot shape, which includes gravity, is developed to quantify the robot shape, friction, and displacement. After validation of the model by experiments, the robot's five actuators are collectively sequenced for inchworm-like forward and backward motion.
\end{abstract}


\section{Introduction\label{subsec:introduction}}

\IEEEPARstart{W}{hile} most soft robots are driven
by pneumatic power \cite{T.2014}, soft robots driven by
piezoelectric actuators may be easier to integrate
\cite{Duggan2019,Cheng2022, 10221022}, less bulky, faster to respond (with driving frequencies up to
 thousands of Hertz \cite{Ji2019,Wu2019}), lighter,
and potentially microsized \cite{Jafferis2019}.

\begin{figure}[tb]
\begin{centering}
\includegraphics[width=0.95\columnwidth]{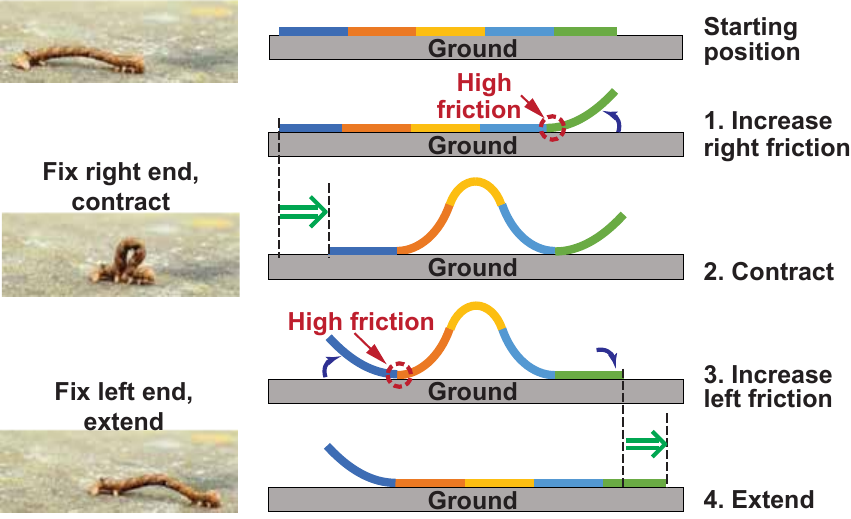}
\par\end{centering}
\caption{Inchworm motion \cite{inchwormpictures} and analogous robot motion of \textquotedbl contract\textquotedbl{}
and \textquotedbl extend\textquotedbl{} cycles in 4 steps. The robot
consists of five thin-film piezoelectric actuators (each shown 
{
in
}
a different
color) on a common substrate. Raising the actuator on one end increases
the friction on that end relative to the opposite end, enabling crawling motion.\label{fig:experiment:robot-motion-design}}
\end{figure}

{
Robotic inchworms are appealing because they in principle can fit through narrow openings \cite{Zheng2022a}, have no rotating parts, and can be made (in our case) by layering different layers on top of one other. By an inchworm, we mean a robot with a
}
motion cycle (Fig. \ref{fig:experiment:robot-motion-design}) that involves first raising its midsection while holding its
front end fixed on a surface, causing its back end to move forward.
In the second step, it releases its front end and fixes its rear end to the surface and extends
its body, causing its front end to move forward. 

{
A number of approaches have been explored to achieve such inchworm-like controllable friction and movement. The driving methods include pneumatic actuators \cite{T.2014, Mosadegh2014, Duggan2019, Das2023}, heat-induced shape changes \cite{Xu2022,Du2022}, embedded magnet pairs controlled by external magnetic field \cite{Joyee2019,Ijaz2020, Maeda2020, Niu}, light-powered liquid crystal elastomer \cite{Ahn2019}. The inchworm-like bio-inspired motion mechanisms include employing friction films on both ends \cite{Calisti2017, Hariri2017,Xie2018, Zheng2022, Das2023}, air suckers at both ends \cite{Du2022}, asymmetric and continuous shape change \cite{Ahn2019}, anisotropic friction pads \cite{Koh2013,Wu2018}, passive or electro-active adhesives \cite{Guo,Huang2021,Murphy2007}, sharp hooks \cite{Duduta2017}, and asymmetric feet \cite{Guo2017,Ijaz2020, Xu2022}.
}

In this paper, we describe a new mechanism to enable inchworm motion by holding one end of such an inchworm robot fixed, and using its shape alone to change the profile of its {contact} force on the ground, and hence control its friction profile. 
This mechanism is validated by analytical models and experiments. 
{
Further, while our work uses only the shift in the contact force profile to cause one end to be fixed while the other moves, the ability to control the contact force on one end vs. the other will clearly be important for other inchworm types which use some kind of ``friction pad'' on one end or the other. 
}

Our soft-robot design comprises five piezoelectric actuators
on a single substrate. The central three actuators cause the
central section to lift off the ground and contract or expand
laterally. This paper shows that the {contact} force of the robot on
the ground
can be shifted between the two ends by lifting the head or tail
off of the ground using the $1^{\text{st}}$ or the $5^{\text{th}}$ actuator. Lifting one end raises the {contact force against the ground} next to it, and reduces the {contact force} at the farther end. 
{
The robot/ground interface thus will have more friction at the lifted end and less friction (enabling it to slide) at the opposite end, 
}
as the maximum static friction force is proportional to the {contact force against the ground}.
The effect is analogous to an inchworm's ability to choose which end has ``sticky feet'' (Fig. \ref{fig:experiment:robot-motion-design}).

{
The main contribution of this paper is to demonstrate both analytically and experimentally that this ``friction asymmetry'' between the two ends of a 2-D piezoelectric soft robot can be electrically controlled by lifting one of the far ends off the ground (using a piezoelectric actuator), enabling a tunable ``inchworm-type motion''. In support of this, the paper has 
{
five}
main sections:}
\begin{enumerate}[label=(\roman*)]
{
\item Section \ref{subsec:experiment:actuator-and-robot-design} introduces the robot design and structure.
}

{
\item 
We develop an analytical model of the robot's shape, including the effects of gravity, on 
how much of the flexible body lifts off the ground,
}
as a function
of the voltages applied to the piezoelectric actuators. The model is compared to experiments, using a robot built from commercially available piezoelectric actuators \cite{wilkie2005nasa} (Section \ref{sec:model:soft-robot-model}).
\item We analytically derive the difference in the {contact force against the ground} on one end vs. the other end as a function of piezoelectric
voltages and provide experimental validation (Section \ref{sec:model:soft-robot-model-ground}).
\item Making use of the resulting asymmetry in friction between the two ends, we
experimentally demonstrate forward and reverse motion of the robot and compare it to the model (Section V).
\item Further, to enable the extrapolation of the results presented here to other experimental conditions, we show how the results would scale to similar robots of different lengths and thicknesses (Section VI).
\end{enumerate}
\section{Robot Design and Construction\label{subsec:experiment:actuator-and-robot-design}}

\begin{figure}[tb]
\begin{centering}
\includegraphics[width=0.95\columnwidth]{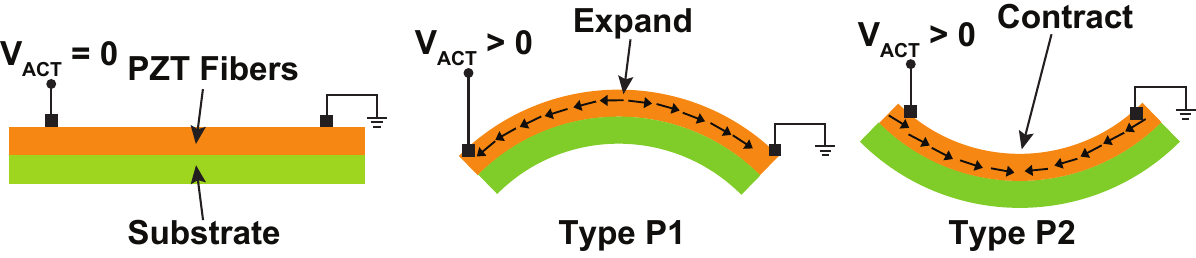}
\par\end{centering}
\caption{
{
Mechanism of bending. Type P1 PZT devices bend concave down with positive applied voltage and type P2 devices bend concave up.\cite{Smartmaterial}
\label{fig:experiment:bending-mechanism}
}
}
\end{figure}

Each actuator of Fig. \ref{fig:experiment:robot-motion-design} is
realized with a 300-\textmu{}m-thick thin-film lead-zirconate-titanate (PZT) device bonded with 100 \textmu m of epoxy to a thin (50-\textmu m) steel substrate. 
Applying a positive voltage causes the piezoelectric material
to expand or contract laterally, resulting in the assembly curling down or up.
(Fig. \ref{fig:experiment:bending-mechanism}). We used commercially
available piezoelectric fiber composite devices \cite{Smartmaterial}.

\begin{figure}[tb]
\begin{centering}
\includegraphics[width=1.0\columnwidth]{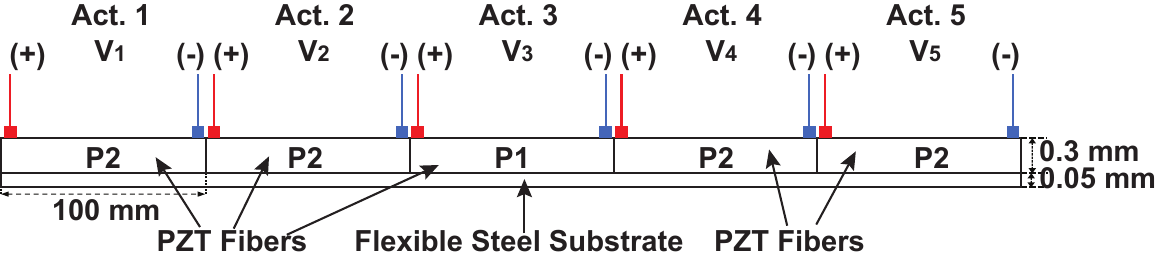}
\par\end{centering}
\caption{
{
Cross-section of a five-actuator soft robot prototype, 500 mm long
and 25 mm wide. Each actuator includes a piezoelectric device made of a
lead zirconate titanate (PZT) fiber composite, controlled by voltage
signals wired from off-robot voltage supplies. All PZT devices are attached to
a common 50-\textmu m-thick steel foil substrate. 
\label{fig:experiment:cross-section}
}
}
\end{figure}

\begin{figure*}[tbh]
\begin{center}
\subfloat[\label{fig:experiment:robot-experimental-setup}]{\begin{centering}
\includegraphics[width=1.9\columnwidth]{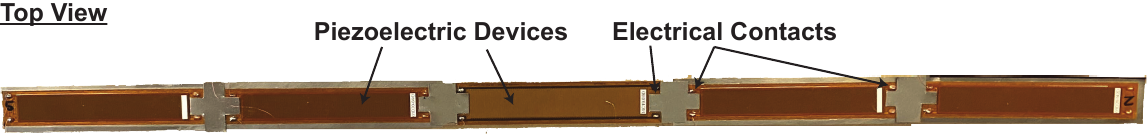}
\par\end{centering}
}

\vfill{}

\subfloat[\label{fig:robot-side-view}]{\begin{centering}
\includegraphics[width=1.9\columnwidth]{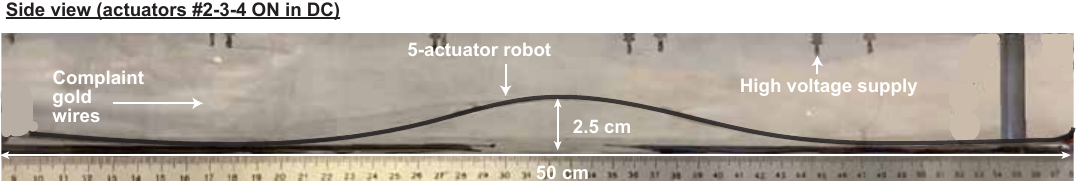}
\par\end{centering}
}

\vfill{}

\subfloat[\label{fig:system-picture}]{\begin{centering}
\includegraphics[width=1.9\columnwidth]{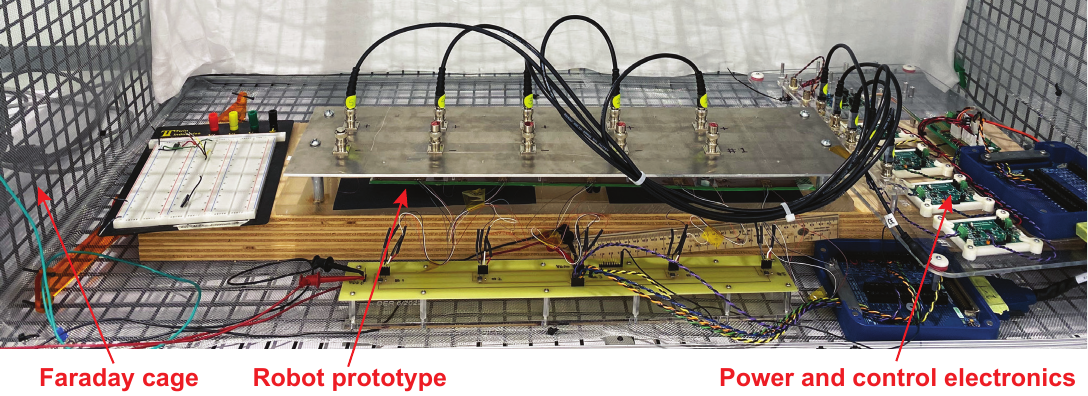}
\par\end{centering}
}
\end{center}

\caption{Robot setup: (a) top view; 
{
(b) side view when the central three actuators are turned on. The 5-actuator
robot is sitting on a rigid acrylic base, wired to high-voltage supplies with
thin gold wires. \cite{Zheng2022}
}
(c) System setup: the system is in a Faraday cage and contains the robot prototype and the power and control electronics.
\label{fig:robot-setup}}
\end{figure*}

Five piezo devices were laminated onto a single steel foil (50 cm long, 2.5 cm wide) to create the robot (Fig. \ref{fig:experiment:cross-section}).
The PZT actuators we used are recommended for use with only one sign of applied voltage, since they degrade if the opposite sign of voltage is applied. Because our high-voltage supplies we used had only a positive output,
in our experiments, we used two kinds of commercially-available piezoelectric composite actuators: one optimized for curling down (when on top of a flexible substrate) with a positive applied voltage (type P1) as in Fig. \ref{fig:experiment:bending-mechanism}, and the second optimized for curling up with a positive applied voltage (type P2), with voltage polarity defined by the markings on each actuator. Thus type P2 actuators (numbers 1,2,4 and 5 in Fig \ref{fig:experiment:cross-section}) only curled up, and type P1 actuator (number 3 in Fig. \ref{fig:experiment:cross-section}) only curled down.

However, for a more general model, the analyses in this paper of the robot shape due to piezoelectric effects assume all actuators have the properties of type P2, with negative voltages used to cause the central actuator (\#3) to curl down. The magnitude of the free strain per volt of the type P1 actuators used in actuator \#3 is a factor of {1.7} smaller than that of the type P2 actuators (0.75 vs 1.3 ppm/V). Thus the experimental applied voltages used on actuator \#3 in our work are {-1.7} times those in the modeling.

Fig. \ref{fig:experiment:robot-experimental-setup} shows the top
view of the robot, and Fig. \ref{fig:robot-side-view}
shows the side view while it is actuated. Thin gold wires
are connected from the solder pads on the actuators to high-voltage
supplies.
The system is put in a Faraday cage and contains the robot prototype as well as power and control electronics (Fig. \ref{fig:system-picture}).

\section{Soft Body Robot Model: Shape\label{sec:model:soft-robot-model}}

Realizing inchworm action relies on alternately raising the
friction of the robot against the ground, between the left and right ends of the robot. Predicting and controlling this friction requires knowledge of the
exact shape of the robot including the effects of gravity on the
shape. 
{
What is also critical is the profile of
}
the vertical force between
the actuator and the ground (which we refer to as \textquotedbl contact
force\textquotedbl{} distribution).

This section develops an analytical static model of the robot and
{contact force} distribution as a function of applied voltages on the
actuators.

\subsection{Previous work}

Control of the robot requires precise analytical modeling. Current methods of modeling a piezoelectric
soft robot have typically employed constant-curvature models \cite{Webster2010,Falkenhahn2015,DellaSantina2020b}
and pseudo-rigid body models \cite{Lobontiu2001,Bandopadhya2010,Li2018,Zheng2022}.
A constant-curvature model treats an actuator (or part of it) as
a perfect arc with some radius. Then a coordinate transformation
can be used to model the kinematics of the robot, a procedure similar to that used for a rigid robot. Alternatively, a pseudo-rigid body model breaks a flexible robot into short rigid links connected by flexible joints.
One can further include the effects of gravity by applying Cosserat
rod theory to a continuum robot in a cantilever case \cite{Jones2009}.

In this paper, we develop an analytical soft-body model for
our robot, which includes gravity and the robot's interaction with the ground. 
Though previously under-investigated, gravity effects are significant and of critical importance because of the robot's elasticity.

\subsection{Soft robot shape modeling}

\subsubsection{Summary of the model}

When voltages are applied to the actuators, some parts of the robot
are flat on the ground, and other parts lift off. In
this section, we develop a model that predicts the shape considering piezoelectricity and gravity. The key property of the model is a self-consistent
approach to determining which part of the robot lifts off the ground.

We developed the model following a ``bottom-up'' approach in three steps: (1) a single actuator clamped on one end; (2) three actuators bending to compress in length like an inchworm;
and (3) a five-actuator inchworm robot, as in Fig.~\ref{fig:experiment:robot-motion-design}.

Our modeling is based on the Euler-Bernoulli small-amplitude model
of displacement $y$ versus the lateral position $x$. In such a model,
for an actuator with a single applied voltage, the shape is determined
by the following physical effects:
\begin{enumerate}[label=(\roman*)]
\item Piezoelectric effect: $\frac{d^{2}y}{dx^{2}}=\gamma V$, where $V$ is the
applied voltage, and $\gamma$ is a constant related with material
properties with a unit of $\text{m}^{-1}\cdot\text{V}^{-1}$. See Appendix \ref{subsec:model:bending-mechanism} for
details.
\item Distributed load $q$ (mass per length): $\frac{d^{4}y}{dx^{4}}=\frac{qg}{EI}$, where
$EI$ is the effective flexural rigidity of our three-layer (tri-morph) structure
(Appendix \ref{subsec:model:bending-mechanism}).
\item Discrete external vertical force $F$: $\frac{d^{3}y}{dx^{3}}=\frac{F}{EI}$.
\item Discrete load mass $m$: discontinuity in $\frac{d^{3}y}{dx^{3}}=-\frac{mg}{EI}$ ($g$ is the gravitational constant).
\end{enumerate}
The model must be applied piecewise because the actuator voltage changes
from one section to another, and because in some sections the actuator
is laying on the ground, whereby the ground applies a vertical force
on the actuator.

The following boundary conditions are applied to the solutions:
\begin{enumerate}[label=(\roman*)]
\item $\frac{dy}{dx}$ must always be continuous, including at all interfaces between
different actuators.
\item Defining the flat ground as $y = 0$, $y$ must be $\geq$ 0 (easily extended in more complex arrangements).
\item The ground must support the weight of the actuator - represented by
a \textquotedbl {contact force}\textquotedbl{} $F_{\text{Ground}}(x)$
acting on the robot, as described later.
\item The total torque on an unconstrained robot about its center of mass
must be zero.
\end{enumerate}

{
This model assumes a small vertical deformation of the robot such that the bending amount must be much less than its length.
}

We now apply this model to increasingly complex shapes, which underlie
the motion of our soft robot. We first analyze a single-actuator cantilever (with one end clamped and the other end free) on the ground, and then proceed to the more complex cases.
\\

\subsubsection{Single actuator with one end clamped parallel to the ground \label{subsec:Single-actuator-clamped}}

\begin{figure}[tb]
\begin{centering}
\includegraphics[width=0.6\columnwidth]{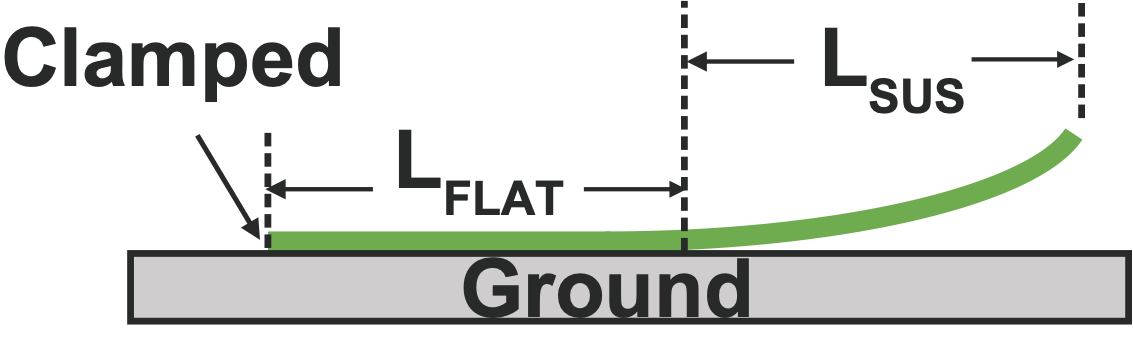}
\par\end{centering}
\caption{One-actuator setup on the ground with the left end clamped. A negative
voltage is applied to make the actuator bend up. One part of
it ($L_{\text{FLAT}}$)
lies flat on the ground due to gravity, the other part ($L_{\text{SUS}}$) is suspended in the air. \label{fig:single-actuator:clamped-setup}}
\end{figure}

Fig. \ref{fig:single-actuator:clamped-setup} shows one actuator placed
on the ground with the left end clamped. When voltage is applied to make the actuator
bend up, part of it ($L_{\text{FLAT}}$) stays flat on the ground due to clamping and gravity
and the other part ($L_{\text{SUS}}$) lifts up. Our goal is to analytically
find the suspended length $L_{\text{SUS}}$.

In the suspended region ($0\leq x\leq L_{\text{SUS}}$), $x=0$ is defined at the point that the robot starts to lift off the ground.
{
The shape of a beam clamped on one end with a distributed gravitational load \cite{Bauchau2009} is well known, as well as the quadratic shape of an unloaded (no gravity) piezoelectric actuator clamped at one end \cite{Weinberg1999}. 
}
By superposition, the shape of the piezoelectric actuator subjected to gravity and clamped on one end (x = 0) is then:
\begin{equation}
y(x)=\frac{1}{2}\gamma Vx^{2}-\frac{qg}{24EI}x^{2}\left(x^{2}-4L_{\text{SUS}}x+6L_{\text{SUS}}^{2}\right)\label{eq:single-actuator:ysum-formula}
\end{equation}

\begin{figure}[tb]
\begin{centering}
\includegraphics[width=0.6\columnwidth]{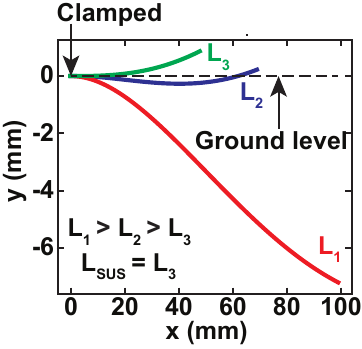}
\par\end{centering}
\caption{Plot of shape for an actuator curling up, with the left end clamped
at $y=0$ and $\frac{dy}{dx}=0$. The suspended length $L_{\text{SUS}}$ would be the largest
$L$ ($L_{3}$) that make the result physical, i.e. all $y(x)>0$.
\label{fig:single-actuator:self-consistent}}
\end{figure}

A self-consistent approach (Fig. \ref{fig:single-actuator:self-consistent})
is proposed to solve $L_{\text{SUS}}$: for small $L_{\text{SUS}}$,
we easily see that $\frac{dy}{dx}$ is always $\geq$ 0, so the suspended
region is indeed suspended. For large $L_{\text{SUS}},$ $y$
might become negative at some $x$, but this is not consistent with our
assumption of the actuator being off the ground. Thus, for any applied voltage,
only a finite length lifts off the ground. The region to the left
of the suspended (``lift-off'') section has $y=0$ and $\frac{dy}{dx}=0$.

\begin{figure}[tb]
\begin{centering}
\includegraphics[width=0.95\columnwidth]{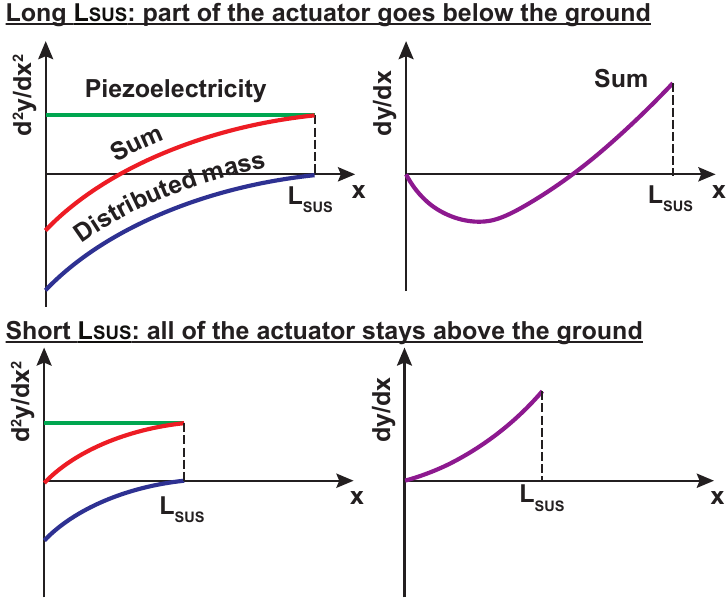}
\par\end{centering}
\caption{Second and first derivatives of the shape caused by piezoelectricity,
gravity, and their sum for different suspended lengths. \label{fig:single-actuator:second-derivative}}
\end{figure}

Because we assume the slope is zero on the left end of the actuator, we examine the second derivative at the left end to determine $L_{\text{SUS}}$.
Fig. \ref{fig:single-actuator:second-derivative} shows the second
derivative components caused by piezoelectricity, gravity, and their
sum. Since the piezoelectricity contribution to
$\frac{d^{2}y}{dx^{2}}$, defined as
$\frac{d^{2}y_{\text{piezo}}}{dx^{2}}$, is
constant and the gravity contribution $\frac{d^{2}y_{\text{weight}}}{dx^{2}}$ is monotonically
increasing, when a ``trial'' lift-off length $L_{\text{SUS}}$ is too long,
the second derivative of the sum $\frac{d^{2}y}{dx^{2}}$ is negative at $x=0$.
So $\frac{dy}{dx}$ becomes negative when $x>0$. Since $y|_{x=0}=0$,
$y$ goes negative as well. This result is not physical, as we
assumed $y>0$ everywhere due to the ground level. 

When reducing the lift-off
length, $\frac{d^{2}y}{dx^{2}}$ increases. When $\frac{d^{2}y}{dx^{2}}|_{x=0}$
reaches $0$, $\frac{dy}{dx}$ becomes positive when $x>0$, and the actuator
is always above ground. This means that $L_{\text{SUS}}$ is the
solution of $\frac{d^{2}y}{dx^{2}}|_{x=0}=0$. Taking the second derivative of Eq. (\ref{eq:single-actuator:ysum-formula})
and setting it equal to 0 at $x = 0$ ($x=0$ is the point that the actuator starts to lift off), one finds:
\begin{equation}
L_{\text{SUS}}=\sqrt{\frac{2EI}{qg}\gamma V}.\label{eq:single-actuator:effective-length-formula}
\end{equation}

Substituting for $L_{\text{SUS}}$ from Eq. (\ref{eq:single-actuator:effective-length-formula})
into Eq. (\ref{eq:single-actuator:ysum-formula}), we have $y(x)$
analytically. For the suspended portion:

\begin{equation}
y(x)=-\frac{qg}{24EI}x^{4}+\frac{\sqrt{2}}{6}\sqrt{\frac{qg}{EI}\gamma V}x^{3},
\label{eq:single-actuator:ysum-result}
\end{equation}

while for the flat portion, we have $y(x)=0$.

We should note that the suspended length $L_{\text{SUS}}$ cannot exceed the actuator length $L$. We need to limit $L_{\text{SUS}}$ at $L$, if the result from Eq. (\ref{eq:single-actuator:effective-length-formula}) exceeds $L$. 
{
In this case, instead of Eq.(\ref{eq:single-actuator:ysum-result}), Eq. (\ref{eq:single-actuator:ysum-formula}) becomes:

\begin{equation}
y(x)=\frac{1}{2}\gamma Vx^{2}-\frac{qg}{24EI}x^{2}\left(x^{2}-4Lx+6L^{2}\right)
\label{eq:single-actuator:ysum-result-lsus-limit}
\end{equation}
}

\begin{figure}[tb]
\begin{centering}
\includegraphics[width=0.9\columnwidth]{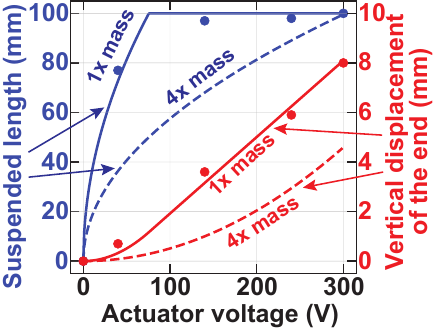}
\par\end{centering}
\caption{
{
The suspended length $L_{\text{SUS}}$ (blue) and displacement (red)
of the free end, versus the driving voltage for a single actuator with
one end clamped parallel to the ground. Model (solid lines) and experimental
measurements (points) show close agreement. Dashed lines represent the modeling when mass is increased by a factor of 4.
}
\label{fig:single-suspended-length-model-experiment}}
\end{figure}

Fig. \ref{fig:single-suspended-length-model-experiment} plots $L_{\text{SUS}}$
and displacement at the lifted end versus actuator voltage, predicted from the model and measured from experiments. The parameter values
are listed in Appendix \ref{subsec:input-material-parameters}. 
Model and experiments show good agreement without any fitting parameters. {
Moreover, we recently began experiments to add onboard batteries, control, and high-voltage circuitry \cite{Cheng2022,Zheng2022b,Cheng2023, 10221022}. This can easily increase the mass of the robot by a factor of four. Assuming the mass is uniformly distributed, we used the above approach to find the effect of the extra mass. The results (dashed lines in Fig. \ref{fig:single-suspended-length-model-experiment}) show such a mass is expected to reduce the vertical displacement by a factor of $\sim$2. This illustrates the importance of including gravity in the models. 
}
\\
\subsubsection{Five-actuator robot -- the middle three actuators\label{subsec:Three-actuators-curl}}

\begin{figure}[tb]
\begin{centering}
\includegraphics[width=0.6\columnwidth]{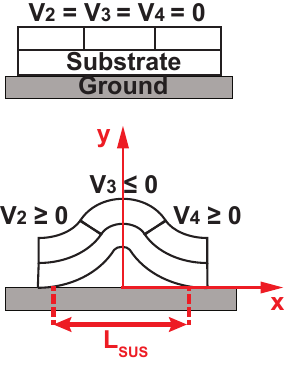}
\par\end{centering}
\caption{Three-actuator assembly with voltages applied such that the center bends concave down and the wings concave up. The projected length of the curved assembly is shorter than its flat length. The central portion of length $L_{\text{SUS}}$ is suspended.
\label{fig:three-actuator:cross-section}}
\end{figure}

Next, we consider how the model predicts the shape of the middle
three actuators (actuators \#2 -- \#4) when placed on the ground and curling up in a symmetric fashion ($V_{\text{2}}=V_{\text{4}}\geq 0$,
$V_{\text{3}}\leq 0$).
Fig. \ref{fig:three-actuator:cross-section}
shows a schematic cross section of the assembly in flat (top) and in bending (bottom) conditions. The left and right ends remain on the ground. We define the lateral length of the total suspended section as $L_{\text{SUS}}$.

In this central suspended section, where there is no force on the robot from the ground, combining piezoelectricity with the Euler-Bernoulli beam equation, defining $x = 0$ at the middle of Actuator \# 3, one finds:
\begin{equation}
y(x)=-\frac{1}{24}\frac{qg}{EI}x^{4}+\frac{1}{2}a_{2}x^{2}+y_{\text{piezo}}(x)+a_{0},\label{eq:three-actuator-displacement}
\end{equation}

\noindent where $y_{\text{piezo}}(x)$ is a function (defined below) which results from piezoelectricity. In the region of actuator \#3 ($-\frac{L}{2} < x < \frac{L}{2}$), where the piezo effect bends the structure concave down:
\begin{equation}
y_{\text{piezo}}(x)=-\frac{1}{2}\gamma|V_{\text{3}}|x^{2},\left(-\frac{L}{2}\leq x\leq\frac{L}{2}\right).
\label{eq:three-actuator:Y-formula-1}
\end{equation}

In the regions of actuators \#2 and \#4 ($\frac{L}{2} < |x| < \frac{3L}{2}$),

\begin{equation}
\begin{aligned}y_{\text{piezo}}(x)= & \begin{cases}
\frac{1}{2}\gamma|V_{\text{4}}|x^{2}-\frac{1}{2}\gamma\left(|V_{\text{3}}|+|V_{4}|\right)L\cdot x \\
+\frac{1}{8}\gamma\left(|V_{\text{3}}|+|V_{\text{4}}|\right)L^{2},\\
\left(\frac{L}{2}\leq x\leq\frac{3L}{2}\right)\\
\frac{1}{2}\gamma|V_{\text{4}}|x^{2}+\frac{1}{2}\gamma\left(|V_{\text{3}}|+|V_{4}|\right)L\cdot x\\
+\frac{1}{8}\gamma\left(|V_{\text{3}}|+|V_{\text{4}}|\right)L^{2},\\
\left(-\frac{3L}{2}\leq x\leq-\frac{L}{2}\right).
\end{cases}\end{aligned}
\label{eq:three-actuator:Y-formula-2}
\end{equation}
The constants $a_{0}$ and $a_{2}$ of Eq. (\ref{eq:three-actuator-displacement}) are obtained from the boundary
conditions $y|_{x=\frac{L_{\text{SUS}}}{2}}=0$ and $\frac{dy}{dx}|_{x=\frac{L_{\text{SUS}}}{2}}=0$:

\begin{equation}
a_{0}=\frac{1}{384}\frac{qg}{EI}L_{\text{SUS}}^{4}+\frac{1}{4}L_{\text{SUS}}\frac{dy_{\text{piezo}}}{dx}|_{x=\frac{L_{\text{SUS}}}{2}}-y_{\text{piezo}}|_{x=\frac{L_{\text{SUS}}}{2}})\label{eq:three-actuator:a0}
\end{equation}

\begin{equation}
a_{2}=\frac{1}{24}\frac{qg}{EI}L_{\text{SUS}}^{2}-\frac{2\frac{dy_{\text{piezo}}}{dx}|_{x=\frac{L_{\text{SUS}}}{2}}}{L_{\text{SUS}}}\label{eq:three-actuator:a2}
\end{equation}

Similar to the analysis in Section \ref{subsec:Single-actuator-clamped}
for a single actuator, $L_{\text{SUS}}$ satisfies $\frac{d^{2}y}{dx^{2}}|_{x=\frac{L_{\text{SUS}}}{2}}=0$. Therefore,
if $L_{\text{SUS}}\geq L$,

\begin{equation}
L_{\text{SUS}}=\sqrt[3]{12\gamma\left(|V_{3}|+|V_{4}|\right)L\frac{EI}{qg}}.\label{eq:three-actuator:suspended-length}
\end{equation}
{
Because we only have 3 actuators, $L_{SUS}$ has a maximum of $3L$.

By substituting Eqs. (\ref{eq:three-actuator:Y-formula-1}) -- (\ref{eq:three-actuator:a2}) into Eq. (\ref{eq:three-actuator-displacement}), one finds: 
}
\begin{equation}
y(0)=-\frac{1}{384}\frac{qg}{EI}L_{\text{SUS}}^{4}+\frac{1}{8}\gamma\left(|V_{3}|+|V_{4}|\right)L(L_{\text{SUS}}-L).\label{eq:three-actuator:origin-displacement}
\end{equation}

{
Physically, $y(0)$ must be greater or equal to 0 for a valid solution, giving a minimum value for applied voltages to have a central section of the robot off the ground. 
For our parameters, the minimum voltage is $\sim$30~V for the ``1x mass'' case.
}

The actuators' shape is derived by substituting $L_{\text{SUS}}$ into (\ref{eq:three-actuator-displacement}),
combining Eqs. (\ref{eq:three-actuator:Y-formula-1}), (\ref{eq:three-actuator:Y-formula-2}) and (\ref{eq:three-actuator:a0}).

\begin{figure}[tb]
\begin{centering}
\includegraphics[width=0.9\columnwidth]{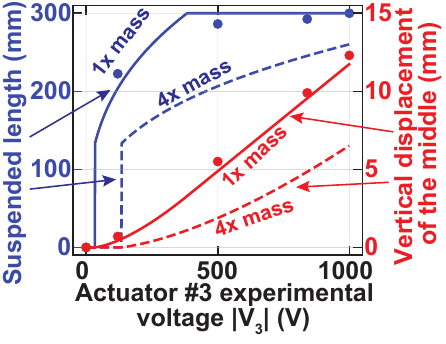}
\par\end{centering}
\caption{
{
The suspended length $L_{\text{SUS}}$ (blue) and the elevation of the center (red) vs the experimental positive voltage applied to actuator \#3 for the three-actuator system,
where experimental voltages $V_{2}=V_{4}=0.3V_{3}$. Model (lines) and experimental measurements
(points) match well. (The model voltages for the actuator \#3 are {-0.58} times the experimental voltages used in the x-axis.) As in Fig.~\ref{fig:single-suspended-length-model-experiment}, the effect of increasing the robot mass by 4 times is also shown. 
}
\label{fig:three-suspended-length-model-experiment}}
\end{figure}

Fig. \ref{fig:three-suspended-length-model-experiment} plots $L_{\text{SUS}}$
and displacement at the midpoint predicted from the model and measured from experiments, as a function of $V_{3}$, and $V_{2}=V_{4}=0.3V_{3}$. {(The voltage ratio comes from the ratio of the maximum recommended voltages for the P2 and P1 actuators. \cite{Smartmaterial}).} (The model voltages for the actuator \#3 are {-0.58} times the experimental voltages used in the x-axis.) As in the case of a single actuator, model and experiments match well.
{
Increasing mass by a factor of 4 is expected to reduce the vertical displacement by a factor of 2, again illustrating the importance of including gravity in the models. 
}

{
As the center of the robot lifts off the ground when voltages are applied, so that the two ends should come closer to one another, one might ask if friction could limit the sliding of the two ends towards each other. Consider three attached actuators, suspended except at the extreme ends. The actuators weigh $\sim$3 grams each for a total of 10 g, for a frictional force $\sim$0.1 Newton, assuming a (large) friction coefficient equal to 1. The “blocking force” of our actuators (what is required to prevent piezoelectric contraction or expansion) ranges from 70-150 N (from the datasheets). Thus, as confirmed by experiment, friction should have little effect on the final piezoelectric shape when the suspended height is much less than the suspended length.
}

\subsubsection{Five-actuator robot\label{subsec:Five-actuator-robot}}

\begin{figure}[tb]
\begin{centering}
\includegraphics[width=0.6\columnwidth]{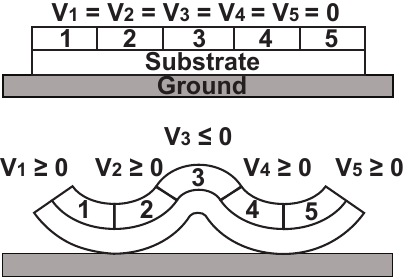}
\par\end{centering}
\caption{The five-actuator robot prototype on the ground. Applying voltages to elements 2, 3 and 4 lifts the center, and makes the projected length of the robot shrink.\label{fig: five-actuator-robot-model:cross-section}}
\end{figure}

Fig. \ref{fig: five-actuator-robot-model:cross-section} shows a schematic cross-section
sketch of a five-actuator robot, corresponding to the arrangement introduced in Figs. \ref{fig:experiment:cross-section} and \ref{fig:robot-setup}. Actuators \#1 to \#5 have applied
voltages $V_{1}$ to $V_{5}$. $V_{1}\geq0$,
$V_{5}\geq0$, $V_{2}=V_{4}\geq0$, and $V_{3}\leq0$.

We first \emph{analytically} examine two cases relevant to robot motion:
\begin{enumerate}[label=(\roman*)]
\item When the voltages applied are low, the interfaces between actuators
\#1 and \#2, and \#4 and \#5 have a flat section, so that we separate the analysis
of the robot into three parts: Section \ref{subsec:Single-actuator-clamped} above
applies to actuator \#1 and \#5, and Section \ref{subsec:Three-actuators-curl}
applies to actuators \#2, \#3, and \#4.
\item When the voltages applied are high enough, the flat regions near the interfaces of actuators \#1 and \#2, and \#4 and \#5, shrink to zero length, so that the robot touches the ground at only two single points. Separate sections of the robot cannot be analyzed independently as described just above, and we need to find where these points are, which we define as $x_{L}$ and $x_{R}$.
{
$x=0$ is defined as the middle point of the robot. 
}
The equation for the shape of the robot depends on location with respect to this $x_{L}$ and $x_{R}$:

{
\begin{equation}
\begin{aligned}y(x)= & \begin{cases}
y_{L}(x) & -\frac{5}{2}L<x\leq x_{L}\\
y_{M}(x) & x_{L}<x\leq x_{R}\\
y_{R}(x) & x_{R}<x\leq \frac{5}{2}L
\end{cases}\end{aligned}
\end{equation}
}

These three shapes all are described by the same form (a 4th-order polynomial), but the coefficients of the terms depend on the applied voltages in each section and the relevant boundary conditions.

\begin{equation}
\begin{aligned}\left(\begin{array}{c}
y_{L}(x)\\
y_{M}(x)\\
y_{R}(x)
\end{array}\right)= & \left(\begin{array}{c}
a_{01}\\
a_{02}\\
a_{03}
\end{array}\right)+\left(\begin{array}{c}
a_{11}\\
a_{12}\\
a_{13}
\end{array}\right)x\\
 & +\frac{1}{2}\left(\begin{array}{c}
a_{21}\\
a_{22}\\
a_{23}
\end{array}\right)x^{2}+\frac{1}{6}\left(\begin{array}{c}
a_{31}\\
a_{32}\\
a_{33}
\end{array}\right)x^{3}\\
 & -\frac{1}{24}\frac{qg}{EI}x^{4}+y_{\text{piezo}}(x)
\end{aligned}
\end{equation}

\end{enumerate}

The twelve parameters $a_{ij}$, $x_{L}$, and $x_{R}$ depend on the applied voltages. They can be found numerically by using the following fourteen boundary conditions: $y_{L}(x_{L})=y_{M}(x_{L})=0$;
$y'_{L}(x_{L})=y'_{M}(x_{L})=0$; 
$y''_{L}(x_{L}) = y''_{M}(x_{L});$
$y_{M}(x_{R})=y_{R}(x_{R})=0$;
$y'_{M}(x_{R})=y'_{R}(x_{R})=0$; 
$y''_{M}(x_{R})=y''_{R}(x_{R})$;
{
$y''(-\frac{5}{2}L)=y_{\text{piezo}}''(-\frac{5}{2}L)$, 
$y'''(-\frac{5}{2}L)=0$;
$y''(\frac{5}{2}L)=y_{\text{piezo}}''(\frac{5}{2}L)$, 
$y'''(\frac{5}{2}L)=0$.
}

\begin{figure}[tb]
\begin{centering}
\includegraphics[height=1.5 in]{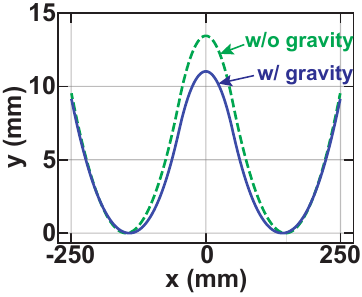}
\par\end{centering}
\caption{
{
Numerical solution of the model for robot shape when $V_{1}=300$ V, $V_{2}=300$
V, $V_{3}={-580}$ V (corresponding to 1000 V in the experiment), $V_{4}=300$ V, and $V_{5}=300$ V with and without gravity. With gravity, the single points that touch the ground are at $x=-141.9$ mm and $x=141.9$ mm. Without gravity, they are -148.1 mm and 148.1 mm.
}
\label{fig:robot-numerical-solution}}
\end{figure}

Fig. \ref{fig:robot-numerical-solution} shows the numerically calculated
solution for $V_{1}=300$ V, $V_{2}=300$ V, $V_{3}={-580}$ V (which would correspond to 1000 V in the experiment), $V_{4}=300$
V, and $V_{5}=300$ V 
{
with and without gravity. Note that with gravity, the ground contact points are not at the actuator junctions themselves but only slightly ($\sim$8 mm) inside the actuators' junctions 1/2 and 3/4. 

If smaller actuator voltages are applied, a ``flat spot'' inside and possibly including the actuator 1/2 and 3/4 interfaces will result, and the lift-off points of the central section will be inside the 1/2 and 3/4 interfaces. Then one can exactly calculate the shape of the suspended hump of the central section, including gravity effects, using Eqn's (\ref{eq:three-actuator-displacement}) -- (\ref{eq:three-actuator:suspended-length}).
}

\begin{figure}
\centering
\subfloat[\label{fig:robot-static-shape-result-1}]{\includegraphics[height=1.8in]{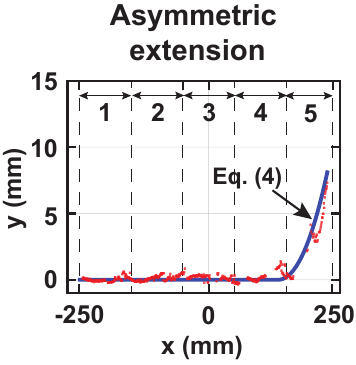}

}\hfill{}\subfloat[\label{fig:robot-static-shape-result-2}]{\includegraphics[height=1.8in]{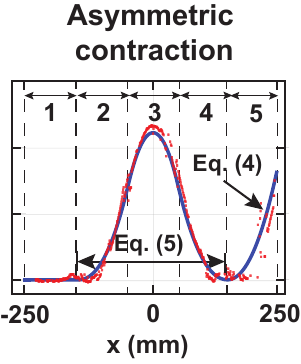}

}
\caption{Robot shapes: five-actuator model vs. experiment. (a) Act \#5 is turned on (numbers 1-5 identify the location of the five actuators); (b) Acts \#2 -- \#5 are turned on for two steps in a movement cycle (steps 1 and 2 of Fig. 1). (Steps 3 and 4 are symmetric to these two).
Voltages when on: $V_{1}=300$ V, $V_{2}=300$ V, $V_{3}= {-580}$ V in modeling ($+1000$ V in experiment with type P1 actuator), $V_{4}=300$
V, $V_{5}=300$ V. Blue lines are modeling result, and red points are points measured from a cross-sectional image of the robot. \label{fig:experiment:robot-shape-model-vs-experiment}}
\end{figure}

\subsubsection{Five-actuator robot: comparison between model and experiment}
We now compare the model solution to experimental measurements for the four steps of the inchworm motion cycle shown schematically in Fig. \ref{fig:experiment:robot-motion-design}.

The applied voltages for actuators when ON are
$V_{1}=300$ V, $V_{2}=300$ V, $V_{3}= {-580}$ V in modeling ($+1000$ V in experiment with type P1 actuator), $V_{4}=300$ V, and $V_{5}=300$ V, (Recall, because actuator \#3 has a different sign and magnitude of piezoelectric coefficient than actuators \#1, \#2, \#4, and \#5, its applied voltage when on was of opposite sign and larger than that of the other actuators, as discussed earlier in Section \ref{subsec:experiment:actuator-and-robot-design}). 

Fig.~\ref{fig:experiment:robot-shape-model-vs-experiment} shows modeled and experimental
robot shapes during two of the four sequential steps in an inchworm movement cycle as described in Fig.~\ref{fig:experiment:robot-motion-design} (with the other two steps being symmetric to these). The voltages are the same as in the previous section. The shape predicted from this model matches closely with experiments. As we describe later, having one end actuator in the air and the other flat leads to asymmetry in friction between the left and right ends of the robot, which is key to its lateral motion. 

We note three omissions in our model, relative to the actual robot design. First, in the actual robot, only $\sim$85\% of the length of each actuator is ``active'' piezoelectric, with the remaining length required for electrical contacts and packaging. Second, while the substrate width is 2.5 cm, the piezoelectric width is only 2.0 cm, which reduces the expected curvature by 8\%, based on straightforward mechanical modeling. Given the inevitable uncertainties in the many other experimental parameters, we ignored these effects. 

{
Third, our 2-D model inherently ignores torsional distortion, i.e., twisting, which could in principle occur when the robot width is narrow and a long central section is suspended off the ground. In experiments with a robot width of 2 cm with a maximum suspended length of 30 cm and height of 2.5 cm (e.g., cross-sectional images such as Fig 4b), any twisting caused a change in the height of one edge of the robot of less a minimum observable amount of $\sim$1 mm.
}
\\

\subsubsection{Lateral motion per inchworm cycle
\label{sec:lateral-motion}}

The amount of forward (or backward) motion from the 4-step cycle in Figs. \ref{fig:experiment:robot-motion-design} and \ref{fig:experiment:robot-shape-model-vs-experiment}
is predicted analytically using our model, by assuming the right end remains fixed in Step 2 and the left end remains fixed in Step 4. Since the substrate
is a steel foil, with a high Young's modulus, its 
\emph{total}
length does not appreciably change during the robot operation. Therefore, the
net lateral movement over 1 cycle is the difference between its lateral length (projection in the x-direction) from Step 1 to 2, or from Step 3 to 4, which we call $L_{x,\text{contract}}$,
where $y(x)$ is given by Eq'ns.
(\ref{eq:three-actuator-displacement}) -- (\ref{eq:three-actuator:suspended-length}), as shown in Fig. \ref{fig:three-actuator:cross-section}. 
One finds:
\begin{equation}
\begin{aligned}L_{x,\text{contract}} & =L_{\text{tot}}-L_{x}\\
 & =\int_{-3L/2}^{3L/2}\left(\sqrt{1+\left(\frac{dy}{dx}\right)^{2}}-1\right)dx\\
 & =2\int_{0}^{L_{\text{SUS}}/2}\left(\sqrt{1+\left(\frac{dy}{dx}\right)^{2}}-1\right)dx
\end{aligned}
\end{equation}
When $|\frac{dy}{dx}|\ll1$, this reduces to: 
\begin{equation}
L_{x,\text{contract}} \approx\int_{0}^{L_{\text{SUS}}/2}\left(\frac{dy}{dx}\right)^{2}dx
\end{equation}
This length difference results from the region of actuators \#2, \#3 and \#4, corresponding to a range in $x$ axis from $-3L/2$ to $+3L/2$ (as defined in Fig. \ref{fig:three-actuator:cross-section}). 

Furthermore, because $|\frac{dy}{dx}|\ll1$ even for the maximum piezoelectric voltages, 
{
when Eq.(\ref{eq:three-actuator:suspended-length}) applies to $L_{\text{SUS}}$,
}
the length contraction $L_{x,\text{contract}}$ is approximated as:
\begin{equation}
\begin{aligned}L_{x,\text{contract}}= & \left(|V_{3}|+|V_{4}|\right)\gamma L^{3}\sqrt[3]{(\frac{qg}{EI})\gamma^{2}L^{2}\left(|V_{3}|+|V_{4}|\right)^{2}}\\
+ & \frac{33\sqrt[3]{12}}{1120}\left(|V_{3}|+|V_{4}|\right)^{2}\gamma^{2}L^{2} \sqrt[3]{(\frac{EI}{qg})\gamma L\left(|V_{3}|+|V_{4}|\right)}\\
- & \frac{1}{12}\gamma^{2}L^{3}\left(|V_{3}|+|V_{4}|\right)^{2}\\
- & \frac{1}{1920}\frac{qg}{EI}\gamma L^{5}\left(|V_{3}|+|V_{4}|\right)
\end{aligned}
\end{equation}

{
When the middle three actuators are all suspended, since the piezoelectric effect is enough to fight the gravity effect, as we noted earlier in Section \ref{subsec:Three-actuators-curl}, $L_{\text{SUS}}$ will be limited at $3L$. In this case, $L_{x,\text{contract}}$ reduces to:

\begin{equation}
\begin{aligned}
L_{x,\text{contract}}= & \frac{1}{18}\left(|V_{3}|+|V_{4}|\right)^{2}\gamma^{2}L^{3}-\frac{(|V_{3}|+|V_{4}|) \gamma q g L^{5}}{12EI}\\
& +\frac{81q^{2}g^{2}L^{7}}{2240 (EI)^{2}}
\label{eq:movement-per-cycle-large-voltage}
\end{aligned}
\end{equation}
When the piezoelectricity is weak, $L_{\text{SUS}} = 0$. Therefore, $L_{x,\text{contract}}=0$.
}
\begin{figure}[tb]
\begin{centering}
\subfloat[\label{fig:model:move-in-amount-vs-V3-a}]{\includegraphics[height=1.50in]{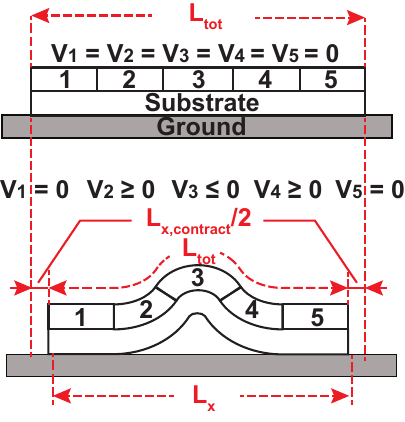}

}\hfill{}\subfloat[\label{fig:model:move-in-amount-vs-V3-b}]{\includegraphics[height=1.50in]{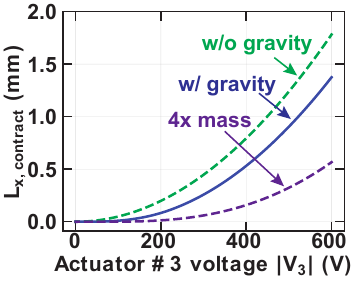}

}
\par\end{centering}
\caption{(a) Schematic view and 
{
(b) modeled lateral motion per inchworm cycle (contraction of length in the x direction in actuators \#2 to \#4) $L_{x,\text{contract}}=L_{\text{tot}}-L_{x}$
versus the magnitude of actuator \#3 voltage $|V_{3}|$, with $V_{2}=V_{4}=0.5|V_{3}|$, with $V_{3}$ as large as {-580} V (corresponding to +1000 V in the experiment). With gravity, $L_{x,\text{contract}}$ and thus lateral movement per cycle are up to 1.1 mm. The cases with mass increased 4 times (purple dashed line) and without gravity (green dashed line) are also plotted.
}
\label{fig:model:move-in-amount-vs-V3}}
\end{figure}

Fig. \ref{fig:model:move-in-amount-vs-V3} shows the lateral motion per cycle $L_{x,\text{contract}}$ as a
function of the magnitude of actuator \#3 voltage $|V_{3}|$ ($V_{3}\leq0$),
while $V_{2}=V_{4}=0.5|V_{3}|$ and $V_{1}=V_{5}=0$ V. $L_{x,\text{contract}}$
increases monotonically with increasing magnitude of $V_{3}$ and
reaches 1.3 mm when $V_{3}=-600$ V (corresponding to 1000 V in the experiment).
{
For comparison, Fig. \ref{fig:model:move-in-amount-vs-V3} also shows the expected contraction per length with no gravity. We see that for our experimental robot, gravity is expected to reduce the motion per cycle by $\sim$22\%. 
Increasing the mass by a factor of 4 is expected to further reduce the motion per cycle by a factor of 3. 
}

\section{Soft body robot model: {contact force against the ground} and friction asymmetry\label{sec:model:soft-robot-model-ground}}
In this section, we use the robot shape to determine the force that the ground exerts on the robot, as a function of location. We then go on to determine the difference in total {contact force against the ground} (and thus the friction experienced) between the left and right sides of the robot, as a function of the applied actuator voltages.

\subsection{{Contact force against the ground} for a single clamped actuator}

We start with the shape of a single actuator with one end clamped parallel
to the ground as derived in Section \ref{subsec:Single-actuator-clamped}, Eq. (\ref{eq:single-actuator:ysum-result}). The distribution of the {contact force} per unit length pushing up on the actuator (``{ground contact force} density'')
is then derived by noting that in the Euler-Bernoulli model, the lateral distribution of the total vertical force on the robot (gravity pulling down minus the ground pushing up) is given by the fourth derivative of the shape. Adding the distributed gravitational force $q\cdot g$ 
(where $q$ is mass/length and $g$ is 9.8 N/kg) then gives the {contact force} pushing up on the robot.

\begin{figure}[tb]
\begin{centering}
\subfloat[\label{fig:single-actuator:ground-force}]{
\includegraphics[width=0.6\columnwidth]{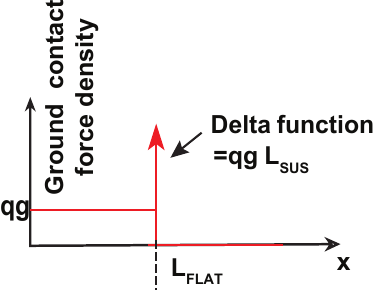}

}

\vfill{}

\subfloat[\label{fig:single-actuator:ground-force-effect-sum}]{
\includegraphics[width=0.6\columnwidth]{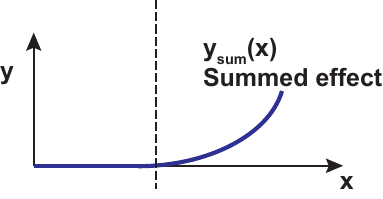}

}

\vfill{}

\subfloat[\label{fig:single-actuator:ground-force-effect-no-ground}]{
\includegraphics[width=0.6\columnwidth]{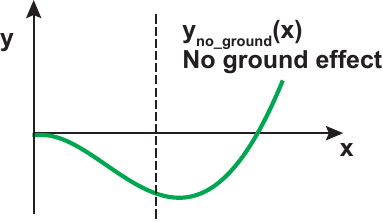}

}

\vfill{}

\subfloat[\label{fig:single-actuator:ground-force-effect-ground}]{
\includegraphics[width=0.6\columnwidth]{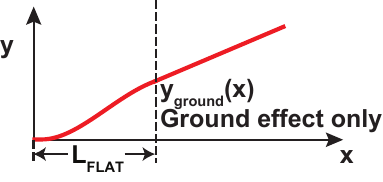}

}

\end{centering}
\caption{
{
(a) {Ground contact force} distribution of a single actuator with its left end clamped
on the ground. (b), (c), (d): 
}
Actuator shape separating the effect of the {contact force}. \label{fig:single-actuator:ground-force-effect}}
\end{figure}

{
Fig. \ref{fig:single-actuator:ground-force} shows
the result (not including the force exerted by the clamp). 
For the section lying flat on the ground, there is no shear force. For an infinitely small segment, as there is no shear force applied on the left end or the right of this segment, gravity and the {contact force against the ground} on this segment need to be balanced. Therefore, the distributed {contact force against the ground} equals the distributed load, which is a constant that equals $qg$ (load per length times gravitational constant). However, for the section lifting off the ground, the {contact force} is zero, since the robot and the ground are not in contact. For the point where “lift-off” begins, the shear force on the flat (left) side is zero (no bending), but the suspended (right) part has the gravity of the whole suspended part acting on it. Therefore, the {contact force} on the “lift-off” point would be the gravitational force of the suspended part, and the density is thus a delta function with amplitude equal to the suspended part’s gravity.
}

The {contact force} distribution can also be understood by separating the
gravity and ground-force effects. Fig. \ref{fig:single-actuator:ground-force-effect-sum}
shows the modeled shape of an actuator ($y(x)$) with applied voltage considering all the physical effects, including piezoelectric, gravity, and {contact force} effects (repeated from Fig \ref{fig:single-actuator:self-consistent}, given by Eq. (\ref{eq:single-actuator:ysum-result})).
Fig. 
\ref{fig:single-actuator:ground-force-effect-no-ground}
shows the shape considering only piezoelectric and gravity force effects (i.e., ``No ground effect''), with shape given by:
\begin{equation}
y_{\text{no\_ground}}(x)=-\frac{qg}{24EI}x^{4}+\frac{qgL}{6EI}x^{3}+\left(\frac{1}{2}\kappa-\frac{qgL^{2}}{4EI}\right)x^{2},
\end{equation}
corresponding to a suspended cantilever. Finally, the difference of these two cases, shown in Fig. \ref{fig:single-actuator:ground-force-effect-ground}, then represents the {contact force} only, given by:
\begin{equation}
\begin{aligned}y_{\text{ground}}(x) & =y_{\text{sum}}(x)-y_{\text{no\_ground}}(x)\\
 & =\frac{qg}{24EI}x^{4}-\frac{qgL}{6EI}x^{3}+\left(\frac{qgL^{2}}{4EI}-\frac{1}{2}\kappa\right)x^{2}.
\end{aligned}
\end{equation}
To the right of the point where the actuator lifts up, there is no force on the actuator, and thus its shape is a straight line.

The shear force caused by the {contact force} is proportional to the
third derivative of the displacement $y_{\text{ground}}(x)$:
\begin{equation}
\begin{aligned}F_{\text{shear, ground}}(x) & =-EI\frac{d^{3}y_{\text{ground}}}{dx^{3}}\\
 & =qgL-qgx & (0\leq x\leq L_{\text{FLAT}}),
\end{aligned}
\end{equation}

and for the suspended part $(x>L_{\text{FLAT}})$:

\begin{equation}
\begin{aligned}F_{\text{shear, ground}}(x)=0 & & (x>L_{\text{FLAT}}).\end{aligned}
\end{equation}

Therefore, the distributed {contact force} is:
\begin{equation}
\begin{aligned}f_{\text{ground}}(x)= & \begin{cases}
qg+qgL_{\text{SUS}}\delta(x-L_{\text{FLAT}}) & (0\leq x\leq L_{\text{FLAT}})\\
0 & (x>L_{\text{FLAT}})
\end{cases}\end{aligned}
\end{equation}

\subsection{{Contact force} and friction asymmetry for the robot
\label{subsec:Friction-asymmetry}}
\subsubsection{Theory}
\begin{figure}[tb]
\begin{centering}
\subfloat[\label{fig:five-actuator-robot-model:friction-asymmetry-no}]{
\includegraphics[width=0.9\columnwidth]{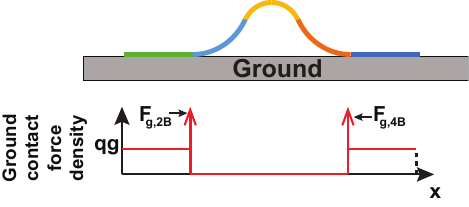}

}

\vfill{}

\subfloat[\label{fig:five-actuator-robot-model:friction-asymmetry-yes}]{
\includegraphics[width=0.9\columnwidth]{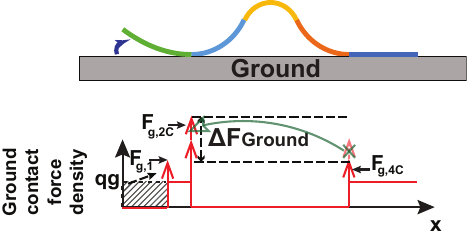}

}
\par\end{centering}
\caption{Mechanism of asymmetry in {contact force against the ground} (and therefore in friction) between the two ends of the robot (``seesaw'' effect): (a) Acts \#2 -- \#4 are turned on; (b) Acts \#1 -- \#4 are turned on.
When the left end rises up transferring vertical {contact force} to the right (the cross-hatched region to impulse function $F_{\text{g,1}}$), torque balance requires that some vertical force is transferred from the right to left side of the robot (indicated schematically the green line).
\label{fig: five-actuator-robot-model:friction-asymmetry}}
\end{figure}

Inchworm motion relies on the ability to alternate the ``stickiness'' between the two ends. In this section, we show how this can be achieved by raising one end of the robot (Actuator \#1 or \#5) off the ground. This raises the static friction at that end, and lowers it at the other end, resulting in the ``friction asymmetry'' necessary for inchworm motion.
Fig. \ref{fig: five-actuator-robot-model:friction-asymmetry} shows
the mechanism of {contact force} and friction asymmetry. We assume friction is positively correlated with {contact force}. So, an increase in {contact force} leads to an increase in friction.
We begin with voltages
$V_{2}$, $V_{3}$, and $V_{4}$ applied, so the middle three actuators
curl up (Fig. \ref{fig: five-actuator-robot-model:friction-asymmetry}a). Then $V_{1}$ is applied to the left actuator, lifting it off the ground (Fig.~\ref{fig: five-actuator-robot-model:friction-asymmetry}b).
Because this section is no longer on the ground, the integrated amount of {contact force} of this section, which was uniformly distributed, now becomes a delta function $F_{g,1}$ at the point where the actuator lifts off the ground. 

This rightward shift of {contact force} induces an imbalance in the torque between the left and right ends of the robot, which previously canceled each other out by symmetry. To maintain a torque balance consistent with the robot shape, an amount $\Delta F_{\text{Ground}}$ of the right {contact force} delta function is transferred to the left side, in a ``seesaw balance'' type of effect. This results in the desired ``friction asymmetry'' between the left and right ends of the robot, as required for ``inchworm'' motion.

We now calculate the magnitude of this $\Delta F_{\text{Ground}}$ (Fig. \ref{fig: five-actuator-robot-model:friction-asymmetry}b). Before raising the actuator, $F_{G,2B}$ and $F_{G,4B}$ each represent half of the mass of the suspended section:
\begin{equation}
F_{g,2B}=F_{g,4B}=\frac{1}{2}qgL_{\text{SUS,MID}}.
\end{equation}
where $L_{\text{SUS,MID}}$ is the suspended length of the middle
three actuators (Eq. (\ref{eq:three-actuator:suspended-length})). 

When actuator \#1 is raised, the {contact} force of the suspended section is transferred to a delta function $F_{g,1}$ (Fig. \ref{fig: five-actuator-robot-model:friction-asymmetry}b), where $F_{g,1}=qgL_{\text{SUS,LEFT}}$.
We now compare the torque on the left and right sides of the middle of the robot about its center, to find $\Delta F_{\text{Ground}}$. On the left side, the torque is:
\begin{equation}
\begin{aligned}
\tau_{L} = & F_{g,1}\left(\frac{5}{2}L-L_{\text{SUS,LEFT}}\right)\\ 
& - qgL_{\text{SUS,LEFT}}\left(\frac{5}{2}L-\frac{1}{2}L_{\text{SUS,LEFT}}\right)\\
& + F_{g,2C}\frac{1}{2}L_{\text{SUS, MID}}.
\end{aligned}
\end{equation}
$F_{g,2C}$ is the discrete {contact force} at the left side of the midsection (Fig. \ref{fig: five-actuator-robot-model:friction-asymmetry}b).
The right side torque is:
\begin{equation}
\begin{aligned}
\tau_{R} = F_{g,4C}\frac{1}{2}L_{\text{SUS, MID}},
\end{aligned}
\end{equation}

\noindent where $F_{g,4C}$ is the impulse {contact force} at the right side of the midsection (Fig. \ref{fig: five-actuator-robot-model:friction-asymmetry}b). Equating these two torques gives:

\begin{equation}
\begin{aligned}F_{g,1}\left(\frac{5}{2}L-L_{\text{SUS,LEFT}}\right)+F_{g,2C}\frac{1}{2}L_{\text{SUS,MID}}\\
=F_{g,4C}\frac{1}{2}L_{\text{SUS,MID}}+qgL_{\text{SUS, LEFT}}\left(\frac{5}{2}L-\frac{1}{2}L_{\text{SUS,LEFT}}\right),
\end{aligned}
\end{equation}
and:
\begin{equation}
F_{g,2C}+F_{g,4C}=qgL_{\text{SUS,MID}}.
\end{equation}

Therefore,

\begin{equation}
F_{g,2C}=\frac{1}{2}qgL_{\text{SUS,MID}}+\frac{qgL_{\text{SUS,LEFT}}^{2}}{2L_{\text{SUS,MID}}}
\end{equation}

\begin{equation}
F_{g,4C}=\frac{1}{2}qgL_{\text{SUS,MID}}-\frac{qgL_{\text{SUS,LEFT}}^{2}}{2L_{\text{SUS,MID}}}
\end{equation}

\begin{equation}
\begin{aligned}\Delta F_{\text{Ground}} & \coloneqq F_{g,4B}-F_{g,4C}\\
 & =\frac{qgL_{\text{SUS,LEFT}}^{2}}{2L_{\text{SUS,MID}}}.
\end{aligned}
\end{equation}
The difference in the total vertical {contact force} between the two sides, which we call the force asymmetry $F_{\text{Asymm}}$, is then:

\begin{equation}
\begin{aligned}F_{\text{Asymm}} & =2\Delta F_{\text{Ground}}\\
 & =\frac{qgL_{\text{SUS,LEFT}}^{2}}{L_{\text{SUS,MID}}}.
\end{aligned}
\end{equation}
As expected, the asymmetry in {contact force} increases as the suspended length of actuator \#1, $L_{\text{SUS,LEFT}}$, increases.

Thus, by raising one end of the robot, the total {contact force} on that end goes up, while going down on the other end, i.e., by lifting up one end of the robot, we increase its friction to the ground compared to the other end, effectively 
{
causing it to stick relative to the other end, as required for inchworm motion.
}

Of interest is the fractional effect - for example the ratio of the difference in the left and right {contact force} vs. the total {contact force} on the right half of the robot ($F_{g,R}$): 
\begin{equation}
\begin{aligned}\frac{F_{\text{Asymm}}}{F_{g,R}} & =\frac{2L_{\text{SUS,LEFT}}^{2}}{L_{\text{SUS,MID}}^{2}+2LL_{\text{SUS, MID}}-L_{\text{SUS,LEFT}}^{2}}.\\
\\
\end{aligned}
\label{eq:model:difference-ratio-ground-force}
\end{equation}

From Eq. (\ref{eq:model:difference-ratio-ground-force}),
$L_{\text{SUS,MID}}=30$ cm and $L_{\text{SUS,LEFT}}=10$ cm as an example, 
$F_{\text{Asymm}}/F_{g,R}=14\%$, implying 
the friction on the left side is 14\% more than on the right side.

The relationship between the ratio and the actuator voltages can be calculated by substituting $L_{\text{SUS, LEFT}}$ and $L_{\text{SUS, MID}}$
from Eqs. (\ref{eq:single-actuator:effective-length-formula})
and (\ref{eq:three-actuator:suspended-length}), when $L_{\text{SUS, LEFT}}\leqslant L$
and $L_{\text{SUS, MID}}\leqslant3L$:
\begin{equation}
\frac{F_{\text{Asymm}}}{F_{g,R}}= \frac{2\gamma V_{1}}{\Big( \sqrt[3]{18\left(\frac{qg}{EI}\right)\left[\gamma L(V_{2}+V_{3})\right]^{2}}
 +\sqrt{2\frac{qg}{EI}\gamma V_{1}}L-\gamma V_{1}\Big)}.
\end{equation}
\\

\subsubsection{Experiments}
Experiments were conducted to validate the {contact force} difference
between the two ends. Five scales (each 9 cm long) were put in a row under the
robot, one per actuator, to
measure the total {contact force} of each section of the robot.

\begin{figure}[tb]
\subfloat[\label{fig:model:ground-force-difference}]{\begin{centering}
\includegraphics[width=0.95\columnwidth]{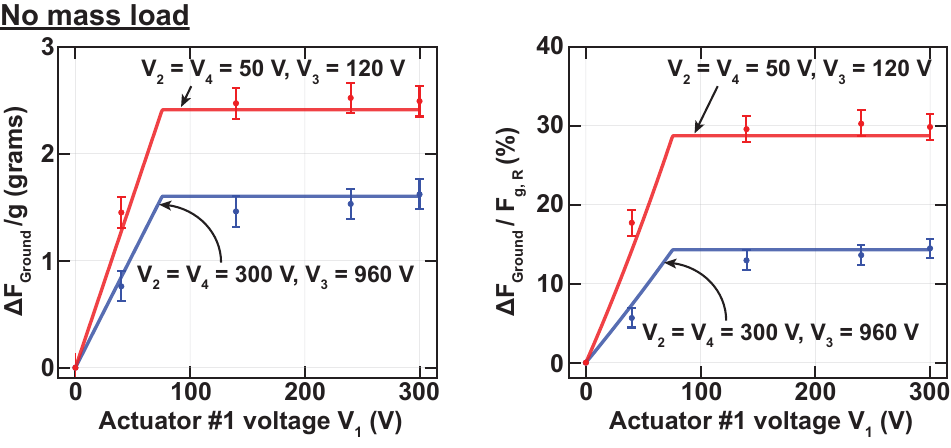}
\par\end{centering}
}

\subfloat[\label{fig:model:ground-force-difference-5g-load}]{\begin{centering}
\includegraphics[width=0.95\columnwidth]{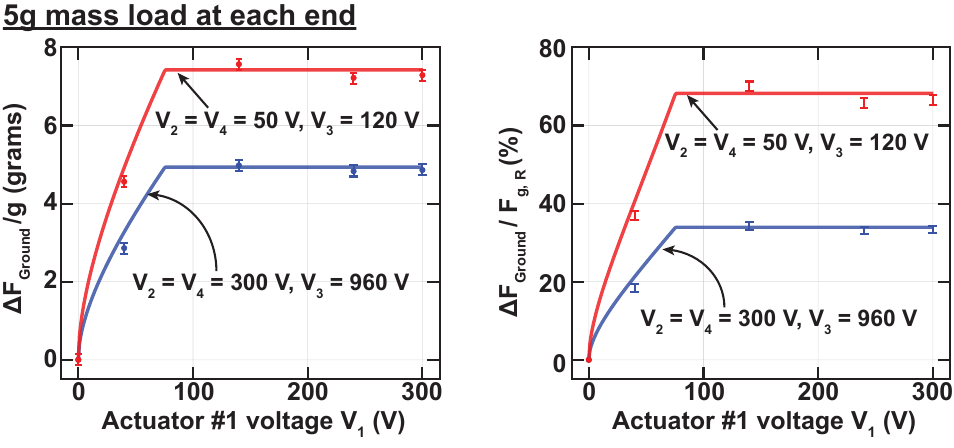}
\par\end{centering}
}

\caption{{Contact force} asymmetry ($F_{\text{Asymm}}$) and the difference
ratio ($F_{\text{Asymm}}/F_{\text{g,R}}$) as a function of
the voltage of the lifting end (actuator \#1), for (a) no mass loads
at the end; (b) 5g mass load at each end. The experiment is run in two different sets of voltages: (i)
$V_{2}=300$ V, $V_{3,\text{exper't.}}=+1000$ V as shown in figure ($V_{3,\text{model}} = {-580}$ V), $V_{4}=300$ V, $V_{5}=0$ V; and
(ii) $V_{2}=50$ V, $V_{3,\text{exper't.}}=+120$ V as shown in figure ($V_{3,\text{model}}=-70$ V), $V_{4}=50$ V, $V_{5}=0$ V.
Error bars represent standard deviations.}
\end{figure}

Fig. \ref{fig:model:ground-force-difference} plots the force asymmetry $F_{\text{Asymm}}$ and the ratio $F_{\text{Asymm}}/F_{g,R}$, from model and experiment as a function of the applied voltages, in two cases: (i) $V_{2}=300$
V, $V_{3}= {-580}$ V in modeling ($+1000$ V in experiment with type P1 actuator), $V_{4}=300$ V, $V_{5}=0$ V; and (ii) $V_{2}=50$
V, $V_{3}=-70$ V ($+120$ V in experiment with type P1 actuator), $V_{4}=50$ V, $V_{5}=0$ V. Error bars represent
standard deviations. Model and experiments match well. For our parameters with real-world actuators, the ground contact
force is as much as 30\% higher on the left end than on the right end.

\subsubsection{Discrete masses to increase friction asymmetry}

Given that surface friction can be highly nonuniform 
{
and/or nonlinear, due to varying surface roughness and other effects}, a difference in {contact force} of 30\% could prove too small to robustly fix one end of the inchworm robot vs. the other.
We now show this asymmetry can be increased by appropriately placed discrete mass loads.
Consider discrete loads $m_{\text{load}}$ placed
on both ends of the robot. With the extra mass, the suspended length of actuator \#1 goes down, but fractionally by a smaller amount than the mass of the suspended length.
This leads to a larger ground-force difference between the two sides:
\begin{equation}
F_{\text{Asymm}}=\frac{qgL_{\text{SUS,LEFT}}^{2}+2 m_{\text{load}}gL_{\text{SUS,LEFT}}}{L_{\text{SUS,MID}}},
\end{equation}
and:
\begin{equation}
\begin{aligned}\frac{F_{\text{Asymm}}}{F_{g,R}}= & \left[2L_{\text{SUS,LEFT}}^{2}+\frac{4m_{\text{load}}gL_{\text{SUS,LEFT}}}{qg}\right]\\
 & /(L_{\text{SUS,MID}}^{2}+2LL_{\text{SUS,MID}}+\frac{2m_{\text{load}}L_{\text{SUS,MID}}}{q}\\
 & -L_{\text{SUS,LEFT}}^{2}-\frac{2m_{\text{load}}L_{\text{SUS,LEFT}}}{q}).
\end{aligned}
\end{equation}

Fig. \ref{fig:model:ground-force-difference-5g-load} shows data which repeats the experiment and model comparison of Fig. \ref{fig:model:ground-force-difference}, but with a 5~g mass on both ends of the robot. It plots the ground
force difference ratio for the same voltages with 5~g mass load at
each end of the robot. The difference is boosted to 70\% from an earlier value of 30\%.

\subsubsection{Experimental validation of adjustable ``friction''}

We now demonstrate that lifting the left end of the robot (without the added-on discrete mass) 
{
increases its friction 
}
compared to the right end, when the total robot length is extended or contracted as in Fig.~\ref{fig:experiment:robot-motion-design}. For these experiments, the robot rests on a smooth plastic (acrylic) sheet. 

\begin{figure}[tb]
\subfloat[\label{fig:experiment:robot-drive-neither-motion-prediction-experiment}]{\begin{centering}
\includegraphics[width=0.95\columnwidth]{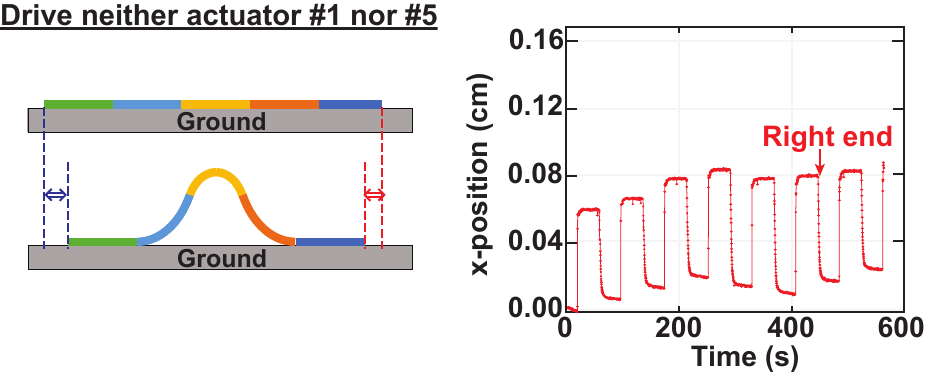}
\par\end{centering}
}

\subfloat[\label{fig:experiment:robot-pull-in-push-out}]{\begin{centering}
\includegraphics[width=0.95\columnwidth]{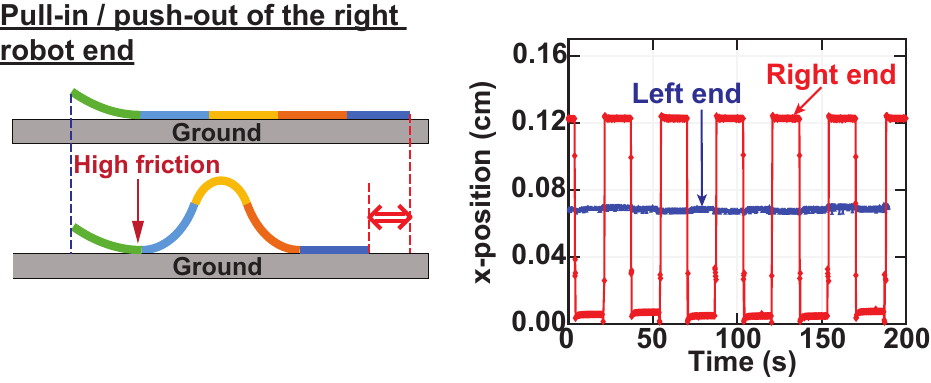}
\par\end{centering}
}
\caption{Validation of motion mechanism for robot lateral motion, in two cases where actuators \#2, \#3, and \#4 are cyclically turned on and off to laterally shrink and expand the central robot section by $\sim$12 mm. (a) The horizontal position of the right end of the robot vs. time, when both
actuators \#1 and \#5 are left off (so they are flat on the ground), and the ends expand and contract symmetrically; (b) Actuator 1 is held in the air to increase friction at the Actuator \#1-\#2 interface, so that the left end is held fixed and all the lateral expansion and contraction occurs on the right end.}
\end{figure}

First, as a control, actuators \#1 and \#5 are not powered, and actuators \#2, \#3 and \#4 are cycled ON (voltages 300 V, 1000 V, 300 V respectively in experiment) and OFF to laterally shrink and extend the robot (Fig. \ref{fig:experiment:robot-drive-neither-motion-prediction-experiment}). Based on the height when the actuators \#2, \#3, \#4 are on, the change in lateral length of the robot is estimated to be 1.2 mm (Section \ref{sec:lateral-motion}). With 
{
similar friction 
}
on either end (a symmetric condition), a change in the right end position of $\sim$0.6 mm is observed over each half-cycle, half of the total length contraction (the left end was observed to symmetrically move by a similar amount in the other direction, data not shown).

The left end of the robot was then held in a raised position ($V_{1}$ = 300 V), and the cycling of the previous figure was repeated (Fig. \ref{fig:experiment:robot-pull-in-push-out}). In this case the left end of the robot is now fixed, moving much less than 0.1 mm, while the right end moves back and forth by the full 1.2 mm. 

This comparison confirms that we can fix one end of the robot to the ground by lifting up the actuator on that end, as required for inchworm motion.

\section{Robot inchworm motion \label{sec:robot-inchworm-motion}}

This section experimentally demonstrates the motion of the inchworm
robot. Results are compared with model predictions for forward motion, backward motion, and the effect of cycling frequency.

\begin{figure}[tb]
\centering
\subfloat[\label{fig:experiment:robot-forward-backward-motion-design}]{\begin{centering}
\includegraphics[width=0.8\columnwidth]{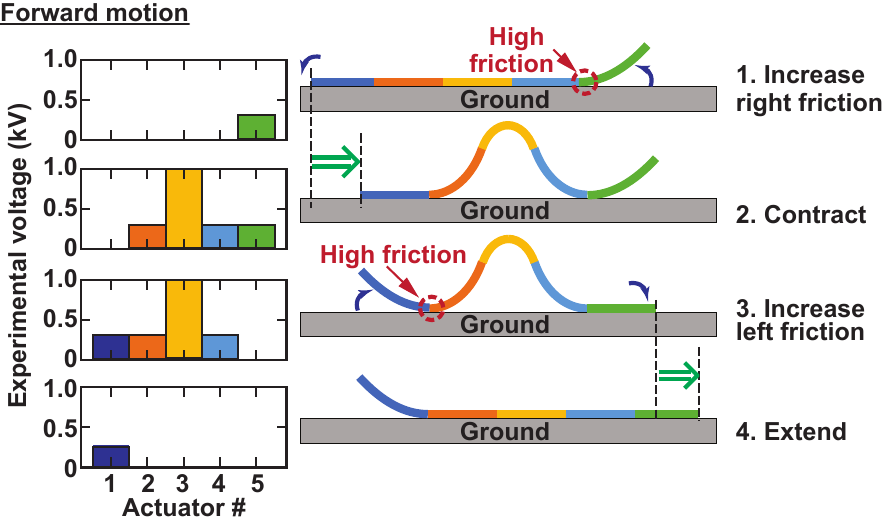}
\par\end{centering}
}

\subfloat[\label{fig:close-up-step-pictures}]{\begin{centering}
\includegraphics[width=0.8\columnwidth]{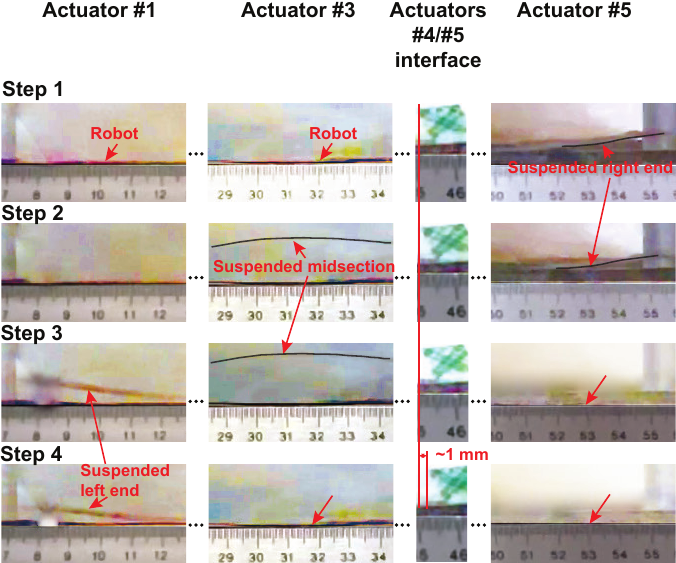}
\par\end{centering}
}

\subfloat[\label{fig:movement-model-experiment}]{\begin{centering}
\begin{tabular}{|>{\centering}p{0.2\columnwidth}|>{\centering}p{0.2\columnwidth}|>{\centering}p{0.2\columnwidth}|>{\centering}p{0.2\columnwidth}|}
\hline 
Step & Left end & Center & Right end\tabularnewline
\hline 
\hline 
Step 1 & & & \tabularnewline
Experiment & 0 mm & 0 mm & 0 mm\tabularnewline
Model & 0 mm & 0 mm & 0 mm\tabularnewline
\hline 
Step 2 & & & \tabularnewline
Experiment & 1.0 mm & 0.5 mm & 0 mm\tabularnewline
Model & 1.2 mm & 0.6 mm & 0 mm\tabularnewline
\hline 
Step 3 & & & \tabularnewline
Experiment & 0 mm & 0 mm & 0 mm\tabularnewline
Model & 0 mm & 0 mm & 0 mm\tabularnewline
\hline 
Step 4 & & & \tabularnewline
Experiment & 0 mm & 0.5 mm & 1.0 mm\tabularnewline
Model & 0 mm & 0.6 mm & 1.2 mm\tabularnewline
\hline 
\end{tabular}
\par\end{centering}
}
\caption{
(a) Robot motion design: one four-step cycle realizing forward motion.
(b) Close-up images of the robot at four positions during each of the four steps of the motion cycle. At the interface between actuators 4 and 5, a lightweight ``flag with an X'' was taped to the robot to be able to discern the lateral motion. The x-scale of images at the interface between actuators 4 and 5 is magnified by 1.3X compared to that at other image locations. During step 4, the robot moves rightward $\sim$1 mm.
{
(c) Experimental and model Lateral movement of left end, center, and right end in one cycle. The experimental error for all the movements is 0.2 mm.
}
}
\end{figure}

Fig. \ref{fig:experiment:robot-forward-backward-motion-design} shows
the design of one cycle of its forward motion with four
steps per cycle, and the voltages supplied for each step. For the forward motion: 
\begin{enumerate}[label=(\roman*)]
\item $V_{1}$ is turned off so that the left end lays flat, and $V_{5}$ is turned on to lift up the right end, increasing its friction to the ground compared to that on the left end. 
\item $V_{2}$ -- $V_{4}$ are turned on to make the middle three actuators
curl up, so the left end moves rightward.
\item $V_{5}$ is turned off to lower the right end and $V_{1}$ is turn on to lift the left end, increasing friction on the left end.
\item The middle section flattens as $V_{2}$ -- $V_{4}$ are turned off, so that the right end moves to the right. 
\item Return to Step 1 to complete a cycle, with the entire inchworm moving to the right. 
\end{enumerate}
The backward motion is similar to the forward motion, but the sequence
of the steps is reversed. 

Fig. \ref{fig:close-up-step-pictures} shows close-up experimental video images of the robot's shape and motion during one cycle. After each cycle, the robot moves rightward (forward) $\sim$1 mm. 
{
Fig. \ref{fig:movement-model-experiment} compares the movement of this cycle with the model prediction for the robot's left end, center, and right end. It again shows a qualitative match between the experiment and the model.
}

\begin{figure}[tb]
\subfloat[\label{fig:experiment:robot-forward-motion-experiment}]{\begin{centering}
\includegraphics[width=0.475\columnwidth]{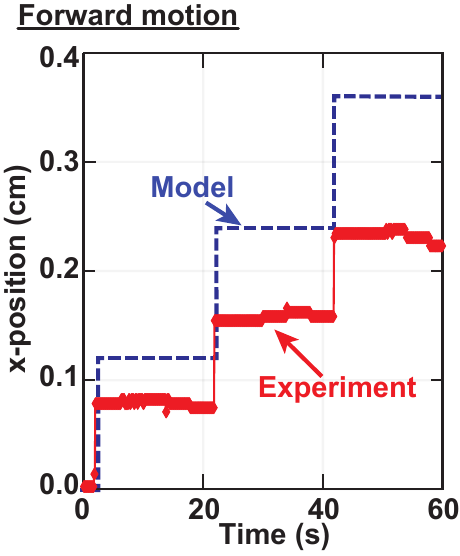}
\par\end{centering}
}\subfloat[\label{fig:experiment:robot-backward-motion-experiment}]{\begin{centering}
\includegraphics[width=0.475\columnwidth]{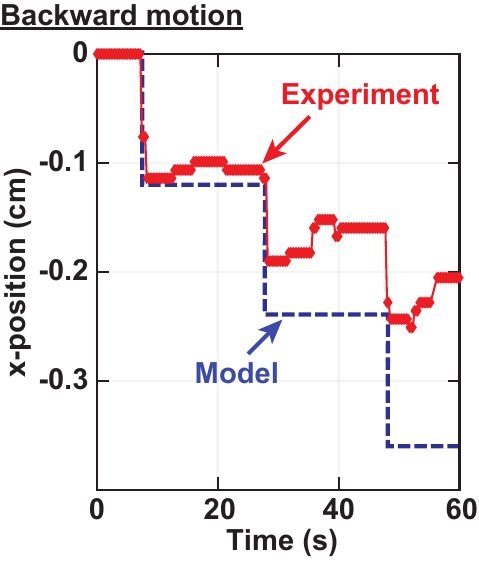}
\par\end{centering}
}

\caption{
{
Inchworm motion of the robot showing the lateral position of its left end (experiment vs. model prediction). (a) The forward motion of the robot using the cycle of Fig. \ref{fig:experiment:robot-forward-backward-motion-design}.
(b) Backward motion. When turned on, the applied voltages in the experiment were $V_{1}=300$ V, $V_{2}=300$ V,
$V_{3}=1000$ V (equivalent to {-580} V in modeling), $V_{4}=300$ V, and $V_{5}=300$ V. The robot moves 0.78 mm on average for each cycle of forward motion,
and 0.66 mm/cycle for backward motion.
}
\label{fig:robot-motion-experiment}
}
\end{figure}

Fig. \ref{fig:robot-motion-experiment}a and b demonstrate
forward and backward motion of the robot, with a time of 5 seconds per step. Both figures show the position of the left end of the robot. Forward motion averages 0.78 mm/cycle and reverse motion 0.66 mm/cycle. These figures demonstrate the ability to fix one end of the robot to the ground vs. the other end (a requirement for ''inchworm'' motion), by raising the actuator on the end to be fixed, thereby increasing the friction on that end of the robot. Note, in the reverse motion there is some evidence of the left end ``backsliding'' during the contraction cycle, accounting for the difference.

{
The model predicts $\sim$1.2 mm per cycle, while the experiment shows a result 30\% smaller. The difference might come from the non-ideal sliding of the ``fixed'' end for some cycles (since Fig. \ref{fig:experiment:robot-pull-in-push-out} shows 1.2 mm movement). Other sources of errors could include uncertainties in the experimental material parameters, such as epoxy thickness between the piezoelectric device and the substrate, non-ideal stresses, or deformations in the as-assembled robot.
}

\begin{figure}[tb]
\centering
\subfloat[\label{fig:experiment:robot-motion-cycle-frequency-scaling}]{\begin{centering}
\includegraphics[width=0.9\columnwidth]{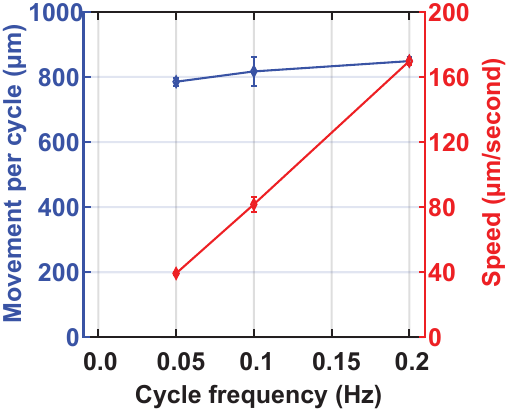}
\par\end{centering}
}

\subfloat[\label{fig:log-movement-frequency}]{\begin{centering}
\includegraphics[width=0.9\columnwidth]{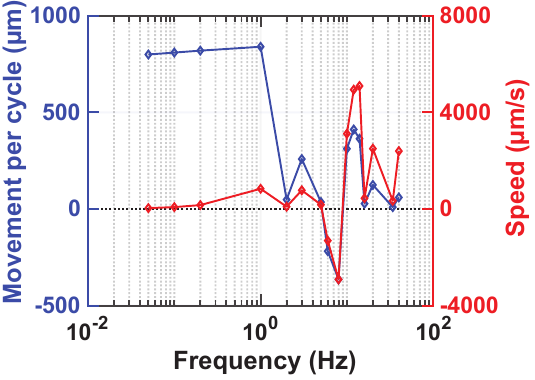}
\par\end{centering}
}\caption{(a) Robot inchworm movement per cycle and speed vs. driving frequency: as the driving cycle period duration changes from 20 s (0.05 Hz) to 5 s (0.2 Hz). The movement per cycle is approximately constant so the average speed increases linearly with driving frequency. {(b)
Robot movement per cycle and speed v.s. driving frequency up to 40 Hz (frequency axis is in a logarithmicscale)\cite{Zheng2022}.}
}
\end{figure}

At low frequencies (> 1~s periods), the lateral motion per cycle was independent of frequency, and a speed of 0.8~mm/s was achieved at 1 Hz (Figs. \ref{fig:experiment:robot-motion-cycle-frequency-scaling} and \ref{fig:log-movement-frequency}). 
When the frequency reaches $\sim$1~Hz, dynamic effects begin to become important, and at higher frequencies the motion slows down and surprisingly the robot can even go backwards. (Fig. \ref{fig:log-movement-frequency})
{
The motion here is no longer an inchworm-like motion.
}
Given that a single actuator (suspended on one end) has a resonance frequency of $\sim$23 Hz (Fig. \ref{fig:single-actuator-frequency-response-experiment}), and that multiple suspended actuators, as in the central section of the robot, will have a much lower resonant frequency, the observed transition of robot motion/cycle at $\sim$1 Hz between the quasi-static {inchworm-motion} regime and a more complicated dynamic motion is reasonable. Modeling of motion in the high-frequency regime is a subject of current investigation.

The energy consumption of the motion is calculated with capacitor charging/discharging models for the actuators, as piezoelectric is an insulator with little static current. For the inchworm motion, all the actuators charge and discharge only once per cycle. Therefore, the energy consumption per movement cycle is:
\begin{equation}
\varepsilon = \sum_{i=1}^{5}{C_{i}V_{i}^{2}}
\end{equation}
By substituting the capacitance ($C_{i}$) of the actuators \cite{Smartmaterial} and our operational voltages ($V_{i}$), the consumed energy per cycle is $\varepsilon = 17$ mJ.

{
\section{Model Implications, Scaling and Limits}
\subsection{Length and thickness scaling}\label{sec:size-scaling-effects}
In this section, with a goal of making a smaller robot, we briefly examine how inchworm movement per cycle ($L_x$) scales with the size of the robot, assuming the length of all elements scales as $S_L$ and the thickness of all layers scale by a factor $S_t$, where $S_t < 1$ for shrinking. (Ignoring aerodynamic effects, as in the rest of the paper, robot width $W$ has no effect.) 
 We assume that the piezoelectric effects are large compared to gravity, as they are in the data presented earlier, so Eq. (\ref{eq:movement-per-cycle-large-voltage}), which is the motion traveled per cycle, becomes:
\begin{equation}
 L_{x,\text{contract}}= \frac{1}{18}\left(V_{3}+V_{4}\right)^{2}\gamma^{2}L^{3}
 \label{eq:movement-per-cycle-simple}
\end{equation}
Assuming type P1 actuators, which depend on the lateral electric field in the piezoelectric layer due to the d$_{33}$ coefficient and have many interdigitated fingers, changing the length changes the number of fingers, but does not change the maximum lateral electric field or maximum voltage in each section. Thus $\alpha$ (free-standing piezoelectric expansion per volt) is independent of $S_{L}$ and $S_{t}$. Because $E_{1}$ (Young's modulus of the piezoelectric) also does not depend on voltage or thickness, and the second moment of area per unit width $I$ scales as ${S_t}^3$, the curvature per voltage (Eq. (\ref{eq:curvature})) $\gamma = \alpha \left(z_{1}E_{1}h_{1}/{EI}\right)$ scales as $1 / {S_t}$. 

Thus the distance travelled per cycle $L_{x,\text{contract}}$ scales as $S_{L}^{3}/S_{t}^{2}$. Thus if the thickness are scaled as 3/2 power of the lengths, the robot motion per cycle is independent of changing its length. (Assuming type P2 actuators leads to the same result.)

{
For example, for a length scaling factor $S_{L}$ of 0.1 (meaning shrinking from the experimental 500 mm to 50 mm) and a thickness scaling factor of $S_{t}$ of $(0.1)^{1.5} = 0.03$, with the same piezoelectric electric field, the robot speed would be unchanged at $\sim$1 mm/s.
}


{
\subsection{Routes toward higher speed}

Other inchworm robots have shown speeds up to 5 mm/s using mechanisms of pneumatic-driven fixing \cite{Liu2022} and embedded magnets \cite{Maeda2020}, respectively. We briefly mention several routes towards higher speed for piezoelectric robots of our general design. First, stronger piezoelectrics (more bending) would increase the motion per cycle. While our commercial piezoelectrics have a $d_{33}$ of 460 pC/N, in more advanced piezoelectric $d_{33}$ can reach 2400 pC/N \cite{Tanaka2009}, 2820 pC/N \cite{Cao2002}, and even 6300 pC/N \cite{Zhong2017}, over 10X larger than those of the actuators used in our work. In principle, this would lead to a 10X increase in $\alpha$ (free-standing strain per volt) and thus a 10X increase in $\gamma$ (curvature per volt), and thus an increase in movement per cycle (Eq. (\ref{eq:movement-per-cycle-simple})) by $10^2$ to over 100 mm/s. An increase in the maximum electric field in the piezoelectric would have a similar effect.

Second, as shown in Fig. \ref{fig:log-movement-frequency}, dynamic effects, which give an observed shape very different than that of the “classic inchworm,” can lead to higher speeds (e.g., 5 mm/s at 14 Hz). A full understanding of such dynamic effects and how they might be exploited is an area of ongoing work. Finally, we note that a similar robot in our lab driven at 14 Hz actually jumps completely off the ground (with a cyclical motion at 7 Hz!)\cite{Zheng2022}. How to understand and exploit such a phenomenon for faster motion will require a thorough understanding of nonlinear dynamic effects and damping in the robot, well beyond the scope of this paper. 
}

\subsection{Surface variability or incline}
In Sec. \ref{subsec:Friction-asymmetry}, we analyzed how much {contact force} difference between the two ends we can generate by changing the robot's shape. Without an extra payload, the difference reaches 14\% and with an extra 5 grams on both ends, it reaches 70\%. These numbers describe how much difference in friction coefficient between the two ends (due to a spatially-varying surface roughness, for example) the robot can tolerate for ideal inchworm motion. If the surface is inclined, the robot must not side downhill. Assuming uniform friction and angle of slope $\theta_{\text{slope}}$, the ``no slide'' condition requires $\theta_{\text{slope}} < \arctan \mu$ where $\mu$ is the friction coefficient. Assuming $\mu = 0.7$, this corresponds to a slope of 35$^{\circ}$.
}

\section{Summary}

Previous work incorporating piezoelectrics into robots has generally relied only on a single actuator for each actuation element. These include a jumping cockroach robot with a single actuator \cite{Wu2019}, a robotic bee with flapping wings \cite{Jafferis2019}, and
a robotic fish with a single actuator as its bending tail \cite{Katzschmann2016}. 

In this work, we introduced a new mechanism for robot motion which inherently depends on the coordinated motion of multiple (five) piezoelectric actuators, by adjusting the friction of one end vs. that on the other end of a multi-element linear piezoelectric inchworm - namely a ``seesaw'' effect. When one end is lifted off the ground, torque balance results in a higher ``{contact force}'' and thus friction to the ground on that end of the robot. Cycling this effect from one end to the other increases the friction of one end vs. the other. No extra physical features such as adhesive or high-friction coating were required. Coupled with a cyclic contraction and expansion of the central section due to the central actuators, the inchworm moved forward or backward as desired. 

Second, to guide our experiments, we developed a first-principle analytical soft-body model for the shape of a linear piezoelectric actuator array with different voltages on different
{
actuators
}
. Key novel aspects are the inclusion of the effect of gravity on the shape, and how the ``{contact force}'' is transferred when sections of the robot lift off the ground. This last effect leads to an asymmetry between the two ends, required for inchworm-type motion.

The models yield excellent matches to experiments on three different levels, all without any adjustable parameters. First, the shape of actuators is well predicted in the presence of gravity, including prediction of how much of the robot lifts off the ground for given applied voltages. Second, as one end of the robot lifts off the ground, how the {contact} force of the robot on the ground moves from one end to the other is well predicted. This was critical for the ultimate lateral motion of the inchworm. Third, motion in both forward and backward directions of 0.1 cm/cycle was predicted.

{
The robot should be well suited for exploring environments with small vertical clearance ($\sim$1-2 cm). While the mechanics of swimming would be different than those discussed here, a similar multi-element thin-film piezoelectric robot 100\% suspended in air \cite{Jafferis2011} was seen to “swim” forward near a surface when a traveling wave was applied. With suitable electrical insulation, the robot of this paper, with much stronger piezoelectrics than the polymer piezoelectrics in \cite{Jafferis2011}, might be adapted for swimming in water.
}

Ongoing work includes faster cycling times for increased speed, modeling system dynamics, and 
{
integration of batteries, sensors, high-voltage electronics, and a Bluetooth microcontroller directly onto the robot for tetherless operation \cite{Cheng2022,Zheng2022b,Cheng2023, 10221022, Aygun2019, Aygun2019a, Sturm_2019, Zheng2019}.
}

\appendix

\subsection{Notations and definitions}

\subsubsection{Input material parameters\label{subsec:input-material-parameters}}

The material parameters used in the model (Table \ref{tab:input-material-parameters})
are taken from the datasheet of the actuator manufacturer, except the bonding
epoxy thickness $h_{2}$, which is measured experimentally.

\begin{table}[tbh]
\caption{Input material parameters \label{tab:input-material-parameters}}

\begin{tabular}{|>{\centering}p{0.2\columnwidth}|>{\centering}p{0.3\columnwidth}|>{\centering}p{0.2\columnwidth}|>{\centering}p{0.1\columnwidth}|}
\hline 
Symbol & Description & Value & Measured (M) / Specified (S)\tabularnewline
\hline 
\hline 
$E_{1}$ & Young's modulus of the PZT device & 30 GPa & S\tabularnewline
\hline 
$E_{2}$ & Young's modulus of the bonding epoxy & 1.5 GPa & S\tabularnewline
\hline 
$E_{3}$ & Young's modulus of the substrate & 203 GPa & S\tabularnewline
\hline 
$\rho_{1}$ & PZT device density & 3.2 g/$\text{cm}^{3}$ & S\tabularnewline
\hline 
$\rho_{2}$ & Epoxy density & 1.1 g/$\text{cm}^{3}$ & S\tabularnewline
\hline 
$\rho_{3}$ & Steel substrate density & 7.9 g/$\text{cm}^{3}$ & S\tabularnewline
\hline 
$h_{1}$ & Thickness of the PZT device & 300 \textmu m & S\tabularnewline
\hline 
$h_{2}$ & Thickness of the bonding epoxy & 93 \textmu m & M\tabularnewline
\hline 
$h_{3}$ & Thickness of the substrate & 50.8 \textmu m & S\tabularnewline
\hline 
$L$ & Length of a single actuator & 10 cm & S\tabularnewline
\hline 
$\alpha$ & Free standing strain per Volt & 1.3 ppm/V for type P1 actuators and 0.75 ppm/V for type P2 actuators & S\tabularnewline
\hline 
\end{tabular}
\end{table}

\subsubsection{Other notations of the model}

Other notations of the model are shown in Table \ref{tab:other-notations-model}.

\begin{table}[tbh]
\caption{Other notations of the model \label{tab:other-notations-model}}

\begin{tabular*}{0.95\columnwidth}{@{\extracolsep{\fill}}|>{\centering}p{0.2\columnwidth}|>{\centering}p{0.65\columnwidth}|}
\hline 
Symbol & Definition\tabularnewline
\hline 
\hline 
$q$ & Distributed mass of the robot per unit length\tabularnewline
\hline 
$R$ & Bending radius of a single actuator\tabularnewline
\hline 
$\kappa$ & Bending curvature of a single actuator\tabularnewline
\hline 
$z_{i}$ & Position of the centerline w.r.t. the neutral axis for the i-th layer\tabularnewline
\hline 
$I_{i}$ & Second moment of area of the i-th layer w.r.t. its centerline\tabularnewline
\hline 
$\epsilon_{1}$ & Magnitude of the electric field in the PZT layer\tabularnewline
\hline 
$EI$ & Effective flexural rigidity of the trimorph structure\tabularnewline
\hline 
$\gamma$ & Bending curvature per unit voltage of the trimorph structure\tabularnewline
\hline 
$L_{\text{FLAT}}$ & Length of the flat part of the clamped single actuator\tabularnewline
\hline 
$L_{\text{SUS}}$ & Length of the suspended part of a single actuator setup\tabularnewline
\hline 
$y_{\text{piezo}}(x)$, $y_{\text{weight}}(x)$ & Displacement caused by piezoelectric effects and gravity\tabularnewline
\hline 
$y_{\text{sum}}(x)$ & Displacement summing up all the effects (piezoelectricity, gravity, and {contact force against the ground})\tabularnewline
\hline 
$y_{\text{no\_ground}}(x)$ & Displacement if there is no {contact force}\tabularnewline
\hline 
$y_{\text{ground}}(x)$ & Displacement due to {contact force}\tabularnewline
\hline 
$F_{\text{shear, ground}}(x)$ & Shear force due to {contact force}\tabularnewline
\hline 
$f_{\text{ground}}(x)$ & Distributed {contact force}\tabularnewline
\hline 
$\kappa(x)$ & $\gamma V(x)$, $V(x)$ is the voltage applied to each actuator\tabularnewline
\hline 
$\theta(x)$ & $\int_{0}^{x}\kappa(x')dx'$\tabularnewline
\hline 
$y_{\text{piezo}}(x)$ & $\int_{0}^{x}\theta(x')dx'$\tabularnewline
\hline 
$F_{g,1}$ & Discrete {contact force} on the lift-off point of actuator \#1\tabularnewline
\hline 
$F_{g,2B}$ & Discrete {contact force} on the lift-off point of actuator \#2 in Fig. \ref{fig:five-actuator-robot-model:friction-asymmetry-no}\tabularnewline
\hline 
$F_{g,4B}$ & Discrete {contact force} on the lift-off point of actuator \#4 in Fig. \ref{fig:five-actuator-robot-model:friction-asymmetry-no}\tabularnewline
\hline 
$F_{g,2C}$ & Discrete {contact force} on the lift-off point of actuator \#2 in Fig. \ref{fig:five-actuator-robot-model:friction-asymmetry-yes}\tabularnewline
\hline 
$F_{g,4C}$ & Discrete {contact force} on the lift-off point of actuator \#4 in Fig. \ref{fig:five-actuator-robot-model:friction-asymmetry-yes}\tabularnewline
\hline 
$\Delta F_{\text{Ground}}$ & The difference of the {contact force} between the left and the right
parts of the robot\tabularnewline
\hline 
$L_{\text{SUS,LEFT}}$, $L_{\text{SUS,MID}}$ & Lengths of suspended part of actuator \#1 and the middle three actuators\tabularnewline
\hline
$S_{L}$, $S_{t}$ & Length and thickness scaling factor of the robot\tabularnewline
\hline 
\end{tabular*}
\end{table}

\subsection{Bending mechanism of the actuator\label{subsec:model:bending-mechanism}}

A single actuator consists of three layers: the top layer is a PZT device, the middle layer is bonding epoxy, and the bottom
layer is a steel substrate. This section describes the bending mechanism
of such an actuator and its design optimization.

\begin{figure}[tb]
\begin{centering}
\includegraphics[width=0.6\columnwidth]{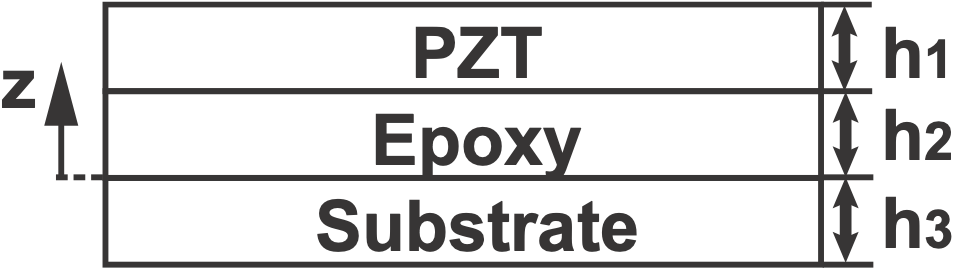}
\par\end{centering}
\caption{Cross-section sketch of the trimorph structure: piezoelectric layer
PZT on the top, bonding epoxy in between and then substrate with thickness
$h_{1}$, $h_{2}$ and $h_{3}$.\label{fig:trimorph:cross-section}}
\end{figure}

Fig. \ref{fig:trimorph:cross-section} shows the cross-section view
of the trilayer structure of an actuator, with PZT thickness $h_{1}$, bonding epoxy
thickness $h_{2}$, and substrate thickness $h_{3}$. When voltage
is applied, the PZT tends to extend, but the substrate tends not to.
The result is that the whole structure bends concave down.
The bending curvature is, referenced w.r.t. the neutral axis \cite{Weinberg1999}:

\begin{equation}
\begin{aligned}
\kappa=\frac{1}{R} & = \alpha V\frac{z_{1}E_{1}h_{1}}{EI}\\
 & =\gamma V,
\label{eq:curvature}
\end{aligned}
\end{equation}

\noindent where $\alpha$ is the free-standing strain per Volt
of the 1st layer. $z_{i}(i=$1, 2, 3) is the position of
the central axis for each layer w.r.t. the neutral axis, $E_{i}$ is
Young's modulus for i-th layer, $EI$
is the flexural rigidity per unit width of the whole structure, $\alpha$
is the free-standing voltage expansion coefficient, $V$ is the applied
voltage, and $\gamma$ is the bending curvature per unit voltage.
We would have:

\begin{equation}
\begin{aligned}EI & =\sum_{i=1}^{3}E_{i}(I_{i}+h_{i}z_{i}^{2}),\\
\\
\end{aligned}
\end{equation}

\noindent where: 
\begin{equation}
\begin{aligned}I_{i} & =\frac{1}{12}h_{i}^{3}\end{aligned},
\end{equation}

\noindent and the position of the neutral axis is:

\begin{equation}
z_{N}=\frac{\sum_{i}z_{i}E_{i}h_{i}}{\sum_{i}E_{i}h_{i}}.
\end{equation}

\subsection{Characterization and modeling of experimental piezoelectric actuators}

\begin{figure}[tb]
\begin{centering}
\includegraphics[width=0.95\columnwidth]{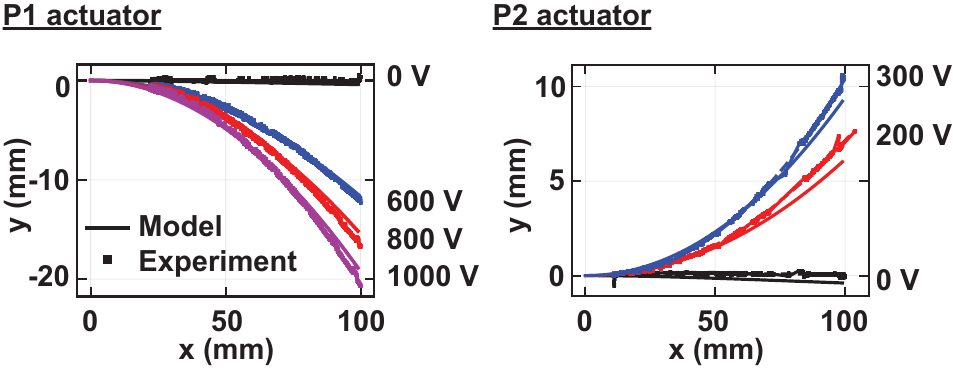}
\par\end{centering}
\caption{Height vs distance for experimental piezoelectric actuators clamped horizontally on their left end. Characterization and modeling of experimental piezoelectric actuators validation for single actuator cantilever settings. A single
actuator is floated in the air, the top surface facing up, clamped
on its left end. Both type P1 actuators (a type of actuator that prefers
to bend down) and type P2 actuators (prefer to bend up) are tested. \label{fig:experiment:model-validation-single-actuator-cantilever}}
\end{figure}

Fig. \ref{fig:experiment:model-validation-single-actuator-cantilever}
shows model validation in single actuator cantilever settings. A single
actuator, as described in Section \ref{subsec:experiment:actuator-and-robot-design},
is floated in the air with its left end clamped. Two kinds of actuators
we used on the robot are tested here: one prefers to bend down
(called ``type P1 actuator''), and the other prefers to bend up (called
``type P2 actuator''). In both scenarios, gravity takes into effect
to pull the actuator down. The voltage ranges that they can take are
also different. Vision sensing described in Section \ref{subsec:Five-actuator-robot}
is used to extract the shape of the actuators for different applied
voltages. Each actuator is 10 cm long. A type P1 actuator can bend down
by about 2 cm with 1000 V supplied from a 5~V-to-1000~V power converter,
and a type P2 actuator can bend up by about 1 cm with 300 V applied. The
experimental results all have good agreements with the modeling results.
{
\begin{figure}[tb]
\begin{centering}
\includegraphics[width=0.6\columnwidth]{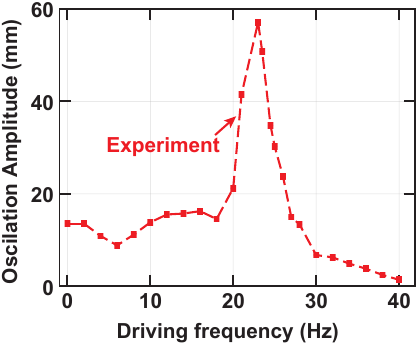}
\par\end{centering}
\caption{
Dynamics single-actuator cantilever showing achieved oscillation amplitude vs. frequency with applied sinusoidal voltages. 
\label{fig:single-actuator-frequency-response-experiment}}
\end{figure}

Fig. \ref{fig:single-actuator-frequency-response-experiment} shows the dynamic behaviors of
an actuator in a cantilever setting, with an applied sinusoidal voltage between 0 V and
1000 V. The resonant frequency is 23 Hz. Note that the operational bandwidth of the piezoelectric device is up to 10 kHz \cite{Smartmaterial}.
}

\subsection{Actuator performance optimization\label{subsec:experiments:actuator-performace-optimization}}

Actuator's bending performance is optimized by tuning the thickness
of the substrate. If the substrate is too thin, it is very soft so
that it would not bend but extends together with the piezoelectric
device. If the substrate is too thick, it is too stiff to be bent by
piezoelectricity. Therefore, there is a sweet spot of substrate thickness
that optimizes the performance. This trade-off needs to be taken into
consideration when designing the actuators.

\begin{figure}[tb]
\begin{centering}
\includegraphics[width=0.6\columnwidth]{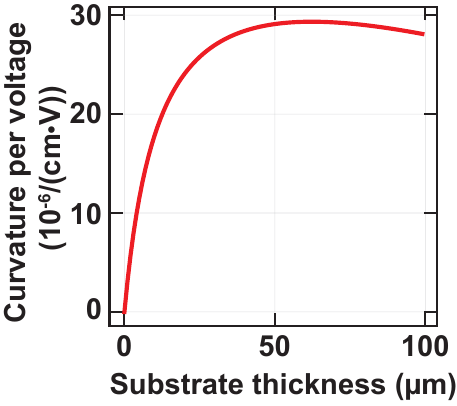}
\par\end{centering}
\caption{Actuator bending performance optimization: bending curvature as a
function of substrate thicknesses.\label{fig:experiment:bending-curvature-vs-substrate-thickness}}
\end{figure}

Fig. \ref{fig:experiment:bending-curvature-vs-substrate-thickness}
shows the bending curvature as a function of substrate thickness.
The thickness of the bonding epoxy is measured experimentally, and
all other material parameters are taken from commercially available
products. The bending curvature reaches its maximum when the substrate
thickness is 65 \textmu m. Therefore, the substrate thickness is picked
to be 50 \textmu m, the nearest one to the optimal point commonly
available.
\bibliographystyle{IEEEtran}
\bibliography{IEEEabrv, mybib}

\begin{thebibliography}{10}
\providecommand{\url}[1]{#1}
\csname url@rmstyle\endcsname
\providecommand{\newblock}{\relax}
\providecommand{\bibinfo}[2]{#2}
\providecommand\BIBentrySTDinterwordspacing{\spaceskip=0pt\relax}
\providecommand\BIBentryALTinterwordstretchfactor{4}
\providecommand\BIBentryALTinterwordspacing{\spaceskip=\fontdimen2\font plus
\BIBentryALTinterwordstretchfactor\fontdimen3\font minus \fontdimen4\font\relax}
\providecommand\BIBforeignlanguage[2]{{%
\expandafter\ifx\csname l@#1\endcsname\relax
\typeout{** WARNING: IEEEtran.bst: No hyphenation pattern has been}%
\typeout{** loaded for the language `#1'. Using the pattern for}%
\typeout{** the default language instead.}%
\else
\language=\csname l@#1\endcsname
\fi
#2}}

\bibitem{T.2014}
\BIBentryALTinterwordspacing
M.~T. Tolley, R.~F. Shepherd, B.~Mosadegh, K.~C. Galloway, M.~Wehner, M.~Karpelson, R.~J. Wood, and G.~M. Whitesides, ``{A Resilient, Untethered Soft Robot},'' \emph{Soft Robotics}, vol.~1, no.~3, pp. 213--223, sep 2014. [Online]. Available: \url{https://www.liebertpub.com/doi/10.1089/soro.2014.0008}
\BIBentrySTDinterwordspacing

\bibitem{Duggan2019}
\BIBentryALTinterwordspacing
T.~Duggan, L.~Horowitz, A.~Ulug, E.~Baker, and K.~Petersen, ``{Inchworm-Inspired Locomotion in Untethered Soft Robots},'' in \emph{2019 2nd IEEE International Conference on Soft Robotics (RoboSoft)}.\hskip 1em plus 0.5em minus 0.4em\relax IEEE, apr 2019, pp. 200--205. [Online]. Available: \url{https://ieeexplore.ieee.org/document/8722716/}
\BIBentrySTDinterwordspacing

\bibitem{Cheng2022}
\BIBentryALTinterwordspacing
H.~Cheng, Z.~Zheng, P.~Kumar, Y.~Chen, and M.~Chen, ``{Hybrid-SoRo: Hybrid Switched Capacitor Power Management Architecture for Multi-Channel Piezoelectric Soft Robot},'' in \emph{2022 IEEE Applied Power Electronics Conference and Exposition (APEC)}.\hskip 1em plus 0.5em minus 0.4em\relax IEEE, mar 2022, pp. 1338--1344. [Online]. Available: \url{https://ieeexplore.ieee.org/document/9773687/}
\BIBentrySTDinterwordspacing

\bibitem{10221022}
H.~Cheng, Z.~Zheng, P.~Kumar, Y.~Chen, J.~Baek, B.~Kim, S.~Wagner, N.~Verma, J.~C. Sturm, and M.~Chen, ``{A Flexible Lightweight 7.4 V Input 300 V to 1500 V Output Power Converter for an Untethered Modular Piezoelectric Soft Robot},'' in \emph{2023 IEEE 24th Workshop on Control and Modeling for Power Electronics (COMPEL)}, 2023, pp. 1--7.

\bibitem{Ji2019}
\BIBentryALTinterwordspacing
X.~Ji, X.~Liu, V.~Cacucciolo, M.~Imboden, Y.~Civet, A.~{El Haitami}, S.~Cantin, Y.~Perriard, and H.~Shea, ``{An autonomous untethered fast soft robotic insect driven by low-voltage dielectric elastomer actuators},'' \emph{Science Robotics}, vol.~4, no.~37, p. eaaz6451, dec 2019. [Online]. Available: \url{https://robotics.sciencemag.org/lookup/doi/10.1126/scirobotics.aaz6451}
\BIBentrySTDinterwordspacing

\bibitem{Wu2019}
\BIBentryALTinterwordspacing
Y.~Wu, J.~K. Yim, J.~Liang, Z.~Shao, M.~Qi, J.~Zhong, Z.~Luo, X.~Yan, M.~Zhang, X.~Wang, R.~S. Fearing, R.~J. Full, and L.~Lin, ``{Insect-scale fast moving and ultrarobust soft robot},'' \emph{Science Robotics}, vol.~4, no.~32, p. eaax1594, jul 2019. [Online]. Available: \url{https://robotics.sciencemag.org/lookup/doi/10.1126/scirobotics.aax1594}
\BIBentrySTDinterwordspacing

\bibitem{Jafferis2019}
\BIBentryALTinterwordspacing
N.~T. Jafferis, E.~F. Helbling, M.~Karpelson, and R.~J. Wood, ``{Untethered flight of an insect-sized flapping-wing microscale aerial vehicle},'' \emph{Nature}, vol. 570, no. 7762, pp. 491--495, jun 2019. [Online]. Available: \url{http://www.nature.com/articles/s41586-019-1322-0}
\BIBentrySTDinterwordspacing

\bibitem{inchwormpictures}
\BIBentryALTinterwordspacing
A.~Dashtpour, ``{https://www.instagram.com/p/COvZDT\_qwac/}.'' [Online]. Available: \url{https://www.instagram.com/p/COvZDT_qwac/}
\BIBentrySTDinterwordspacing

\bibitem{Zheng2022a}
\BIBentryALTinterwordspacing
Z.~Zheng, P.~Kumar, Y.~Chen, H.~Cheng, S.~Wagner, M.~Chen, N.~Verma, and J.~C. Sturm, ``{Model-Based Control of Planar Piezoelectric Inchworm Soft Robot for Crawling in Constrained Environments},'' in \emph{2022 IEEE 5th International Conference on Soft Robotics (RoboSoft)}.\hskip 1em plus 0.5em minus 0.4em\relax IEEE, apr 2022, pp. 693--698. [Online]. Available: \url{https://ieeexplore.ieee.org/document/9762147/}
\BIBentrySTDinterwordspacing

\bibitem{Mosadegh2014}
\BIBentryALTinterwordspacing
B.~Mosadegh, P.~Polygerinos, C.~Keplinger, S.~Wennstedt, R.~F. Shepherd, U.~Gupta, J.~Shim, K.~Bertoldi, C.~J. Walsh, and G.~M. Whitesides, ``{Pneumatic Networks for Soft Robotics that Actuate Rapidly},'' \emph{Advanced Functional Materials}, vol.~24, no.~15, pp. 2163--2170, apr 2014. [Online]. Available: \url{https://onlinelibrary.wiley.com/doi/10.1002/adfm.201303288}
\BIBentrySTDinterwordspacing

\bibitem{Das2023}
\BIBentryALTinterwordspacing
R.~Das, S.~P.~M. Babu, F.~Visentin, S.~Palagi, and B.~Mazzolai, ``{An earthworm-like modular soft robot for locomotion in multi-terrain environments},'' \emph{Scientific Reports}, vol.~13, no.~1, pp. 1--14, 2023. [Online]. Available: \url{https://doi.org/10.1038/s41598-023-28873-w}
\BIBentrySTDinterwordspacing

\bibitem{Xu2022}
L.~Xu, R.~J. Wagner, S.~Liu, Q.~He, T.~Li, W.~Pan, Y.~Feng, H.~Feng, Q.~Meng, X.~Zou, Y.~Fu, X.~Shi, D.~Zhao, J.~Ding, and F.~J. Vernerey, ``{Locomotion of an untethered, worm-inspired soft robot driven by a shape-memory alloy skeleton},'' \emph{Scientific Reports}, vol.~12, no.~1, pp. 1--16, 2022.

\bibitem{Du2022}
T.~Du, L.~Sun, and J.~Wan, ``{A Worm-like Crawling Soft Robot with Pneumatic Actuators Based on Selective Laser Sintering of TPU Powder},'' \emph{Biomimetics}, vol.~7, no.~4, 2022.

\bibitem{Joyee2019}
E.~B. Joyee and Y.~Pan, ``{A fully three-dimensional printed inchworm-inspired soft robot with magnetic actuation},'' \emph{Soft Robotics}, vol.~6, no.~3, pp. 333--345, 2019.

\bibitem{Ijaz2020}
\BIBentryALTinterwordspacing
S.~Ijaz, H.~Li, M.~C. Hoang, C.~S. Kim, D.~Bang, E.~Choi, and J.~O. Park, ``{Magnetically actuated miniature walking soft robot based on chained magnetic microparticles-embedded elastomer},'' \emph{Sensors and Actuators, A: Physical}, vol. 301, p. 111707, 2020. [Online]. Available: \url{https://doi.org/10.1016/j.sna.2019.111707}
\BIBentrySTDinterwordspacing

\bibitem{Maeda2020}
K.~Maeda, H.~Shinoda, and F.~Tsumori, ``{Miniaturization of worm-type soft robot actuated by magnetic field},'' \emph{Japanese Journal of Applied Physics}, vol.~59, no.~SI, 2020.

\bibitem{Niu}
\BIBentryALTinterwordspacing
H.~Niu, R.~Feng, Y.~Xie, B.~Jiang, Y.~Sheng, Y.~Yu, H.~Baoyin, and X.~Zeng, ``{MagWorm: A Biomimetic Magnet Embedded Worm-Like Soft Robot},'' \emph{Soft Robotics}, vol.~8, no.~5, pp. 507--518, oct 2021. [Online]. Available: \url{https://www.liebertpub.com/doi/10.1089/soro.2019.0167}
\BIBentrySTDinterwordspacing

\bibitem{Ahn2019}
C.~Ahn, X.~Liang, and S.~Cai, ``{Bioinspired Design of Light-Powered Crawling, Squeezing, and Jumping Untethered Soft Robot},'' \emph{Advanced Materials Technologies}, vol.~4, no.~7, pp. 1--9, 2019.

\bibitem{Calisti2017}
\BIBentryALTinterwordspacing
M.~Calisti, G.~Picardi, and C.~Laschi, ``{Fundamentals of soft robot locomotion},'' \emph{Journal of The Royal Society Interface}, vol.~14, no. 130, p. 20170101, may 2017. [Online]. Available: \url{https://royalsocietypublishing.org/doi/10.1098/rsif.2017.0101}
\BIBentrySTDinterwordspacing

\bibitem{Hariri2017}
\BIBentryALTinterwordspacing
H.~H. Hariri, G.~S. Soh, S.~Foong, and K.~Wood, ``{Locomotion Study of a Standing Wave Driven Piezoelectric Miniature Robot for Bi-Directional Motion},'' \emph{IEEE Transactions on Robotics}, vol.~33, no.~3, pp. 742--747, jun 2017. [Online]. Available: \url{http://ieeexplore.ieee.org/document/7855815/}
\BIBentrySTDinterwordspacing

\bibitem{Xie2018}
\BIBentryALTinterwordspacing
R.~Xie, M.~Su, Y.~Zhang, M.~Li, H.~Zhu, and Y.~Guan, ``{PISRob: A Pneumatic Soft Robot for Locomoting Like an Inchworm},'' in \emph{2018 IEEE International Conference on Robotics and Automation (ICRA)}.\hskip 1em plus 0.5em minus 0.4em\relax IEEE, may 2018, pp. 3448--3453. [Online]. Available: \url{https://ieeexplore.ieee.org/document/8461189/}
\BIBentrySTDinterwordspacing

\bibitem{Zheng2022}
\BIBentryALTinterwordspacing
Z.~Zheng, P.~Kumar, Y.~Chen, H.~Cheng, S.~Wagner, M.~Chen, N.~Verma, and J.~C. Sturm, ``{Scalable Simulation and Demonstration of Jumping Piezoelectric 2-D Soft Robots},'' in \emph{2022 International Conference on Robotics and Automation (ICRA)}.\hskip 1em plus 0.5em minus 0.4em\relax IEEE, may 2022, pp. 5199--5204. [Online]. Available: \url{https://ieeexplore.ieee.org/document/9811927/}
\BIBentrySTDinterwordspacing

\bibitem{Koh2013}
\BIBentryALTinterwordspacing
J.-S. Koh and K.-J. Cho, ``{Omega-Shaped Inchworm-Inspired Crawling Robot With Large-Index-and-Pitch (LIP) SMA Spring Actuators},'' \emph{IEEE/ASME Transactions on Mechatronics}, vol.~18, no.~2, pp. 419--429, apr 2013. [Online]. Available: \url{http://ieeexplore.ieee.org/document/6269102/}
\BIBentrySTDinterwordspacing

\bibitem{Wu2018}
\BIBentryALTinterwordspacing
Y.~Wu, K.~Y. Ho, K.~Kariya, R.~Xu, W.~Cai, J.~Zhong, Y.~Ma, M.~Zhang, X.~Wang, and L.~Lin, ``{PRE-curved PVDF/PI unimorph structures for biomimic soft crawling actuators},'' in \emph{2018 IEEE Micro Electro Mechanical Systems (MEMS)}, vol. 2018-Janua, no. January.\hskip 1em plus 0.5em minus 0.4em\relax IEEE, jan 2018, pp. 581--584. [Online]. Available: \url{https://ieeexplore.ieee.org/document/8346620/}
\BIBentrySTDinterwordspacing

\bibitem{Guo}
\BIBentryALTinterwordspacing
J.~Guo, C.~Xiang, A.~Conn, and J.~Rossiter, ``{All-Soft Skin-Like Structures for Robotic Locomotion and Transportation},'' \emph{Soft Robotics}, vol.~7, no.~3, pp. 309--320, jun 2020. [Online]. Available: \url{https://www.liebertpub.com/doi/10.1089/soro.2019.0059}
\BIBentrySTDinterwordspacing

\bibitem{Huang2021}
\BIBentryALTinterwordspacing
J.~Huang, Y.~Liu, Y.~Yang, Z.~Zhou, J.~Mao, T.~Wu, J.~Liu, Q.~Cai, C.~Peng, Y.~Xu, B.~Zeng, W.~Luo, G.~Chen, C.~Yuan, and L.~Dai, ``{Electrically programmable adhesive hydrogels for climbing robots},'' \emph{Science Robotics}, vol.~6, no.~53, p. eabe1858, apr 2021. [Online]. Available: \url{https://robotics.sciencemag.org/lookup/doi/10.1126/scirobotics.abe1858}
\BIBentrySTDinterwordspacing

\bibitem{Murphy2007}
\BIBentryALTinterwordspacing
M.~P. Murphy and M.~Sitti, ``{Waalbot: An Agile Small-Scale Wall-Climbing Robot Utilizing Dry Elastomer Adhesives},'' \emph{IEEE/ASME Transactions on Mechatronics}, vol.~12, no.~3, pp. 330--338, jun 2007. [Online]. Available: \url{http://ieeexplore.ieee.org/document/4244394/}
\BIBentrySTDinterwordspacing

\bibitem{Duduta2017}
\BIBentryALTinterwordspacing
M.~Duduta, D.~R. Clarke, and R.~J. Wood, ``{A high speed soft robot based on dielectric elastomer actuators},'' in \emph{2017 IEEE International Conference on Robotics and Automation (ICRA)}.\hskip 1em plus 0.5em minus 0.4em\relax IEEE, may 2017, pp. 4346--4351. [Online]. Available: \url{http://ieeexplore.ieee.org/document/7989501/}
\BIBentrySTDinterwordspacing

\bibitem{Guo2017}
\BIBentryALTinterwordspacing
H.~Guo, J.~Zhang, T.~Wang, Y.~Li, J.~Hong, and Y.~Li, ``{Design and control of an inchworm-inspired soft robot with omega-arching locomotion},'' in \emph{2017 IEEE International Conference on Robotics and Automation (ICRA)}, no.~c.\hskip 1em plus 0.5em minus 0.4em\relax IEEE, may 2017, pp. 4154--4159. [Online]. Available: \url{http://ieeexplore.ieee.org/document/7989477/}
\BIBentrySTDinterwordspacing

\bibitem{wilkie2005nasa}
W.~K. Wilkie, ``{NASA MFC piezocomposites: A development history},'' in \emph{Proceedings of ISMA}, vol. 2014, 2005.

\bibitem{Smartmaterial}
\BIBentryALTinterwordspacing
``{Smart Material Corp. Sarasota, Florida. Part numbers: M-8514-P1, M-8514-P2}.'' [Online]. Available: \url{https://www.smart-material.com/MFC-product-mainV2.html}
\BIBentrySTDinterwordspacing

\bibitem{Webster2010}
\BIBentryALTinterwordspacing
R.~J. Webster and B.~A. Jones, ``{Design and Kinematic Modeling of Constant Curvature Continuum Robots: A Review},'' \emph{The International Journal of Robotics Research}, vol.~29, no.~13, pp. 1661--1683, nov 2010. [Online]. Available: \url{http://journals.sagepub.com/doi/10.1177/0278364910368147}
\BIBentrySTDinterwordspacing

\bibitem{Falkenhahn2015}
\BIBentryALTinterwordspacing
V.~Falkenhahn, A.~Hildebrandt, R.~Neumann, and O.~Sawodny, ``{Model-based feedforward position control of constant curvature continuum robots using feedback linearization},'' in \emph{Proceedings - IEEE International Conference on Robotics and Automation}, vol. 2015-June, no. June.\hskip 1em plus 0.5em minus 0.4em\relax IEEE, may 2015, pp. 762--767. [Online]. Available: \url{http://ieeexplore.ieee.org/document/7139264/}
\BIBentrySTDinterwordspacing

\bibitem{DellaSantina2020b}
\BIBentryALTinterwordspacing
C.~{Della Santina}, A.~Bicchi, and D.~Rus, ``{On an Improved State Parametrization for Soft Robots With Piecewise Constant Curvature and Its Use in Model Based Control},'' \emph{IEEE Robotics and Automation Letters}, vol.~5, no.~2, pp. 1001--1008, apr 2020. [Online]. Available: \url{https://ieeexplore.ieee.org/document/8961972/}
\BIBentrySTDinterwordspacing

\bibitem{Lobontiu2001}
\BIBentryALTinterwordspacing
N.~Lobontiu, M.~Goldfarb, and E.~Garcia, ``{A piezoelectric-driven inchworm locomotion device},'' \emph{Mechanism and Machine Theory}, vol.~36, no.~4, pp. 425--443, apr 2001. [Online]. Available: \url{https://linkinghub.elsevier.com/retrieve/pii/S0094114X00000562}
\BIBentrySTDinterwordspacing

\bibitem{Bandopadhya2010}
\BIBentryALTinterwordspacing
D.~Bandopadhya, ``{Derivation of Transfer Function of an IPMC Actuator Based on Pseudo-Rigid Body Model},'' \emph{Journal of Reinforced Plastics and Composites}, vol.~29, no.~3, pp. 372--390, feb 2010. [Online]. Available: \url{http://journals.sagepub.com/doi/10.1177/0731684408097778}
\BIBentrySTDinterwordspacing

\bibitem{Li2018}
\BIBentryALTinterwordspacing
W.-B. Li, W.-M. Zhang, H.-X. Zou, Z.-K. Peng, and G.~Meng, ``{A Fast Rolling Soft Robot Driven by Dielectric Elastomer},'' \emph{IEEE/ASME Transactions on Mechatronics}, vol.~23, no.~4, pp. 1630--1640, aug 2018. [Online]. Available: \url{https://ieeexplore.ieee.org/document/8365835/}
\BIBentrySTDinterwordspacing

\bibitem{Jones2009}
\BIBentryALTinterwordspacing
B.~A. Jones, R.~L. Gray, and K.~Turlapati, ``{Three dimensional statics for continuum robotics},'' in \emph{2009 IEEE/RSJ International Conference on Intelligent Robots and Systems}.\hskip 1em plus 0.5em minus 0.4em\relax IEEE, oct 2009, pp. 2659--2664. [Online]. Available: \url{http://ieeexplore.ieee.org/document/5354199/}
\BIBentrySTDinterwordspacing

\bibitem{Bauchau2009}
\BIBentryALTinterwordspacing
O.~A. Bauchau and J.~I. Craig, ``{Euler-Bernoulli beam theory},'' in \emph{Structural Analysis. Solid Mechanics and Its Applications, vol 163}, 2009, pp. 173--221. [Online]. Available: \url{http://link.springer.com/10.1007/978-90-481-2516-6_5}
\BIBentrySTDinterwordspacing

\bibitem{Weinberg1999}
\BIBentryALTinterwordspacing
M.~Weinberg, ``{Working equations for piezoelectric actuators and sensors},'' \emph{Journal of Microelectromechanical Systems}, vol.~8, no.~4, pp. 529--533, 1999. [Online]. Available: \url{http://ieeexplore.ieee.org/document/809069/}
\BIBentrySTDinterwordspacing

\bibitem{Zheng2022b}
\BIBentryALTinterwordspacing
Z.~Zheng, H.~Cheng, P.~Kumar, S.~Wagner, M.~Chen, N.~Verma, and J.~C. Sturm, ``{Wirelessly-Controlled Untethered Piezoelectric Planar Soft Robot Capable of Bidirectional Crawling and Rotation},'' in \emph{2023 IEEE International Conference on Robotics and Automation (ICRA)}.\hskip 1em plus 0.5em minus 0.4em\relax IEEE, may 2023, pp. 641--647. [Online]. Available: \url{https://ieeexplore.ieee.org/document/10160886/}
\BIBentrySTDinterwordspacing

\bibitem{Cheng2023}
\BIBentryALTinterwordspacing
H.~Cheng, Z.~Zheng, P.~Kumar, W.~Afridi, B.~Kim, S.~Wagner, N.~Verma, J.~C. Sturm, and M.~Chen, ``{eViper: A Scalable Platform for Untethered Modular Soft Robots},'' mar 2023. [Online]. Available: \url{http://arxiv.org/abs/2303.01676}
\BIBentrySTDinterwordspacing

\bibitem{Liu2022}
X.~Liu, M.~Song, Y.~Fang, Y.~Zhao, and C.~Cao, ``{Worm‐Inspired Soft Robots Enable Adaptable Pipeline and Tunnel Inspection},'' \emph{Advanced Intelligent Systems}, vol.~4, no.~1, p. 2100128, 2022.

\bibitem{Tanaka2009}
\BIBentryALTinterwordspacing
D.~Tanaka, T.~Tsukada, M.~Furukawa, S.~Wada, and Y.~Kuroiwa, ``{Thermal Reliability of Alkaline Niobate-Based Lead-Free Piezoelectric Ceramics},'' \emph{Japanese Journal of Applied Physics}, vol.~48, no.~9, p. 09KD08, sep 2009. [Online]. Available: \url{https://iopscience.iop.org/article/10.1143/JJAP.48.09KD08}
\BIBentrySTDinterwordspacing

\bibitem{Cao2002}
\BIBentryALTinterwordspacing
H.~Cao and H.~Luo, ``Elastic, piezoelectric and dielectric properties of pb(mg 1/3 nb 2/3 )o 3 -38\%pbtio 3 single crystal,'' \emph{Ferroelectrics}, vol. 274, no.~1, pp. 309--315, 2002. [Online]. Available: \url{https://doi.org/10.1080/00150190213965}
\BIBentrySTDinterwordspacing

\bibitem{Zhong2017}
\BIBentryALTinterwordspacing
J.~Zhong, Q.~Zhong, X.~Zang, N.~Wu, W.~Li, Y.~Chu, and L.~Lin, ``{Flexible PET/EVA-based piezoelectret generator for energy harvesting in harsh environments},'' \emph{Nano Energy}, vol.~37, no.~1, pp. 268--274, jul 2017. [Online]. Available: \url{https://linkinghub.elsevier.com/retrieve/pii/S2211285517303063}
\BIBentrySTDinterwordspacing

\bibitem{Katzschmann2016}
\BIBentryALTinterwordspacing
R.~K. Katzschmann, A.~D. Marchese, and D.~Rus, ``{Hydraulic Autonomous Soft Robotic Fish for 3D Swimming},'' in \emph{Springer Tracts in Advanced Robotics}.\hskip 1em plus 0.5em minus 0.4em\relax Springer, Cham, 2016, vol. 109, pp. 405--420. [Online]. Available: \url{http://link.springer.com/10.1007/978-3-319-23778-7_27}
\BIBentrySTDinterwordspacing

\bibitem{Jafferis2011}
\BIBentryALTinterwordspacing
N.~T. Jafferis, H.~A. Stone, and J.~C. Sturm, ``{Traveling wave-induced aerodynamic propulsive forces using piezoelectrically deformed substrates},'' \emph{Applied Physics Letters}, vol.~99, no.~11, p. 114102, sep 2011. [Online]. Available: \url{http://aip.scitation.org/doi/10.1063/1.3637635}
\BIBentrySTDinterwordspacing

\bibitem{Aygun2019}
\BIBentryALTinterwordspacing
L.~E. Aygun, P.~Kumar, Z.~Zheng, T.-S. Chen, S.~Wagner, J.~C. Sturm, and N.~Verma, ``{Hybrid LAE-CMOS Force-Sensing System Employing TFT-Based Compressed Sensing for Scalability of Tactile Sensing Skins},'' \emph{IEEE Transactions on Biomedical Circuits and Systems}, vol.~13, no.~6, pp. 1264--1276, dec 2019. [Online]. Available: \url{https://ieeexplore.ieee.org/document/8877761/}
\BIBentrySTDinterwordspacing

\bibitem{Aygun2019a}
L.~Aygun, P.~Kumar, Z.~Zheng, T.-S. Chen, S.~Wagner, J.~Sturm, and N.~Verma, ``{17.3 Hybrid System for Efficient LAE-CMOS Interfacing in Large-Scale Tactile-Sensing Skins via TFT-Based Compressed Sensing},'' in \emph{Digest of Technical Papers - IEEE International Solid-State Circuits Conference}, vol. 2019-Febru, 2019.

\bibitem{Sturm_2019}
\BIBentryALTinterwordspacing
J.~Sturm, Y.~Mehlman, L.~E. Aygun, C.~Wu, Z.~Zheng, P.~Kumar, S.~Wagner, and N.~Verma, ``{(Keynote) Machine Learning and High-Speed Circuitry in Thin Film Transistors for Sensor Interfacing in Hybrid Large-Area Electronic Systems},'' \emph{ECS Transactions}, vol.~92, no.~4, p. 121, jul 2019. [Online]. Available: \url{https://dx.doi.org/10.1149/09204.0121ecst}
\BIBentrySTDinterwordspacing

\bibitem{Zheng2019}
\BIBentryALTinterwordspacing
Z.~Zheng, L.~E. Aygun, Y.~Mehlman, S.~Wagner, N.~Verma, and J.~C. Sturm, ``{Analyzing and Increasing Yield of ZnO Thin-Film Transistors for Large-area Sensing Systems by Preventing Process-Induced Gate Dielectric Breakdown},'' in \emph{2019 Device Research Conference (DRC)}, vol. 2019-June.\hskip 1em plus 0.5em minus 0.4em\relax IEEE, jun 2019, pp. 141--142. [Online]. Available: \url{https://ieeexplore.ieee.org/document/9046406/}
\BIBentrySTDinterwordspacing

\end{thebibliography}

\begin{IEEEbiography}[{\includegraphics[width=1in,height=1.25in,clip,keepaspectratio]{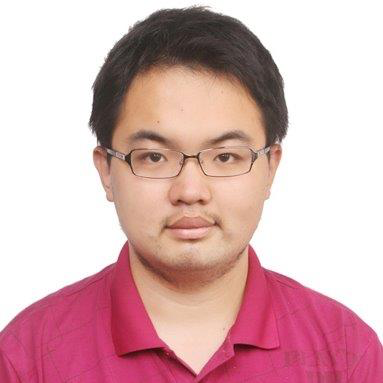}}]{Zhiwu Zheng}(Member, IEEE) received his Ph.D. degree in Electrical and Computer Engineering from Princeton University, Princeton, NJ, USA, in May 2023, and his B.S. degree in Physics from Nanjing University, Nanjing, China, in June, 2016. He is currently a Senior Machine Learning Research Scientist at Analog Devices, Inc.

His current research interests include machine learning and statistical signal processing algorithms, architectures, circuits and devices for advanced sensing systems and robotics. 

Mr. Zheng was the recipient of an ICRA Best Poster Award, an IROS Best Paper Finalist, a RoboSoft Outstanding Poster Award, 2018-2023 Fellowships of the Program in Plasma Science and Technology (PPST) at Princeton University, 2014 Shi Shi-Yuan Scholarship and 2014 Infineon Scholarship, as well as 2013-2016 Fellowships of China National Innovational Undergraduates Program, Nanjing University 2013 Excellent Student, 2013 Scientific Innovation Scholarship, and 2013-2014, 2014-2015, 2015-2016 First-Place Talent Scholar Awards.\end{IEEEbiography}
\vskip -2\baselineskip plus -1fil
\begin{IEEEbiography}[{\includegraphics[width=1in,height=1.25in,clip,keepaspectratio]{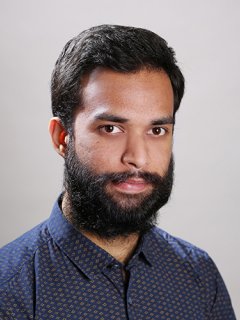}}]{Prakhar Kumar}
 received B.E.(Hons) in Electrical and Electronics Engineering from BITS Pilani, Pilani Campus, India in 2016. 
 
 His current research focuses on developing hybrid sensing systems by bringing together Large Area Electronics (LAE) and CMOS VLSI technologies. More specifically, he is interested in designing technology platforms that can enable machine learning algorithms to carry out inference and state estimation tasks effectively.\end{IEEEbiography}
\vskip -2\baselineskip plus -1fil
 \begin{IEEEbiography}[{\includegraphics[width=1in,height=1.25in,clip,keepaspectratio]{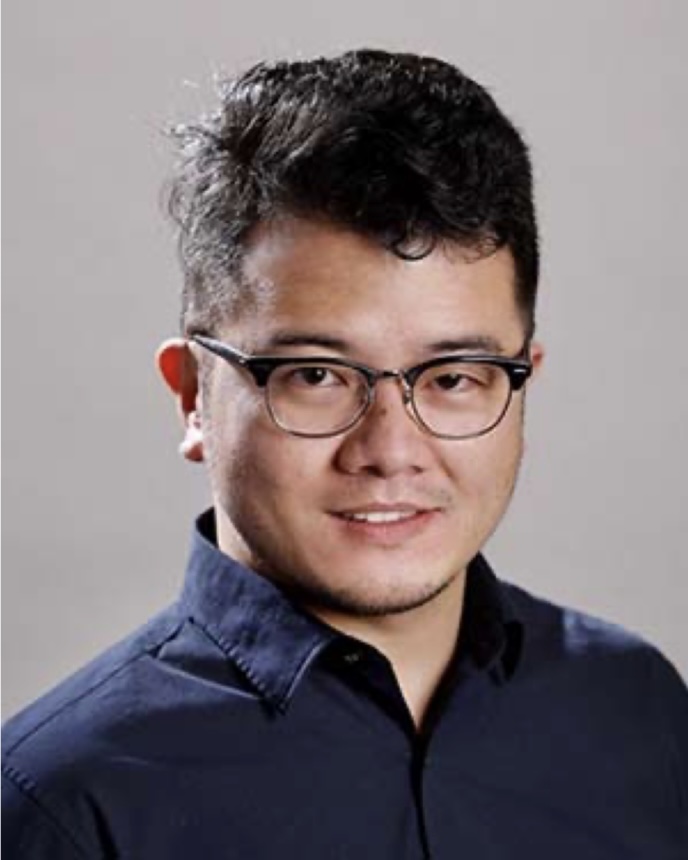}}]{Yenan Chen}(Member, IEEE) received the Honors degree in engineering from Chu Kochen College, Zhejiang University, Hangzhou, China, in 2010, and the bachelor's and Ph.D. degrees in electrical engineering from the College of Electrical Engineering, Zhejiang University, in 2010 and 2018, respectively.,From 2018 to 2021, he was a Postdoctoral Research Associate with the Department of Electrical Engineering, Princeton University, Princeton, NJ, USA. Since December 2021, he has been a Principal Investigator with the ZJU-Hangzhou Global Scientific and Technological Innovation Center, Zhejiang University, where he has been a ZJU100 Professor with the College of Electrical Engineering since July 2023. He holds seven issued Chinese patents. His research interests include power electronic topology, architecture, and control for data center, renewable energy, and transportation.,Dr. Chen was the recipient of two Prize Paper Awards of IEEE Transactions on Power Electronics in 2021 and 2022, the IEEE COMPEL Best Paper Award in 2020, the IEEE Applied Power Electronics Conference and Exposition Outstanding Presentation Award in 2019, and the First Place Award from the Innovation Forum of Princeton University in 2019.\end{IEEEbiography}
\vskip -2\baselineskip plus -1fil
\begin{IEEEbiography}[{\includegraphics[width=1in,height=1.25in,clip,keepaspectratio]{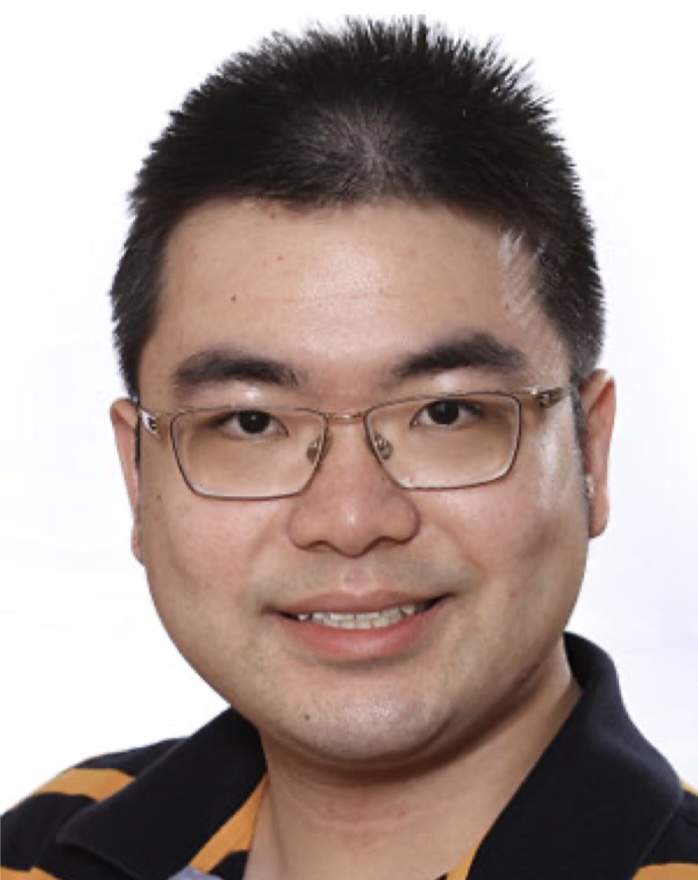}}]{Hsin Cheng} (Student Member, IEEE) received the B.S. degree in engineering from the Department of Electrical Engineering, National Taiwan University, Taipei, Taiwan, in 2020. He is currently working toward the Ph.D. degree with the Department of Electrical and Computer Engineering, Princeton University, Princeton, NJ, USA. His research interests include high-voltage-conversion-ratio modular power architecture, design and mechanism of electroactive soft robots, and their intersections. \end{IEEEbiography}
\vskip -2\baselineskip plus -1fil
\begin{IEEEbiography}[{\includegraphics[width=1in,height=1.25in,clip,keepaspectratio]{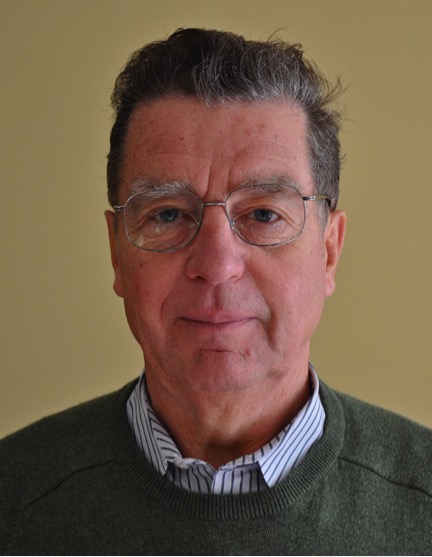}}]{Sigurd Wagner} (Life Fellow, IEEE) is Professor Emeritus of Electrical Engineering and Senior Scholar at Princeton University, where he has been working since 1980.  He received his Ph.D. in physical chemistry in 1968 from the University of Vienna in Austria, came to the US as a post-doc at Ohio State University, worked at Bell Labs from 1970 to 1978, and was the founding chief of the Photovoltaic Research Branch at the Solar Energy Research Institute (now NREL) from 1978 to 1980.  At Princeton he has been developing concepts and applications of large-area electronics, and has been an advocate of laboratory-based learning.

In collaboration with colleagues and students, he has introduced fundamentally new materials, concepts, devices, and structures to thin-film electronics:  The first heterojunction solar cell with a ternary compound semiconductor, CuInSe2, and a quaternary compound, Cu2CdSnS4, among others; the Fermi-level dependence of the density of recombination center in hydrogenated amorphous silicon; microfluidic fundamentals of forming device patterns by printing, and the design and fabrication of microfluidic devices;  foundational experiments on flexible, rollable, elastically stretchable, and conformably shaped electronic surfaces, and their basic design rules and architecture.  At present he is working with his colleagues James C. Sturm and Naveen Verma on the demonstration of complete large-area hybrid thin-film/CMOS-circuit systems for sensing applications.

He is a Life Fellow of the IEEE, a Fellow of the American Physical Society, and a corresponding member of the Austrian Academy of Sciences.  He was an Alexander von Humboldt Foundation Senior Fellow, received the 2009 Nevill Mott Prize, the 2014 ITC 10th Anniversary Prize, and the 2017 MRS David Turnbull Lectureship.  A plaque in Princeton’s Electrical Engineering undergraduate laboratory honors his engagement in lab education.\end{IEEEbiography}
\vskip -2\baselineskip plus -1fil
\begin{IEEEbiography}[{\includegraphics[width=1in,height=1.25in,clip,keepaspectratio]{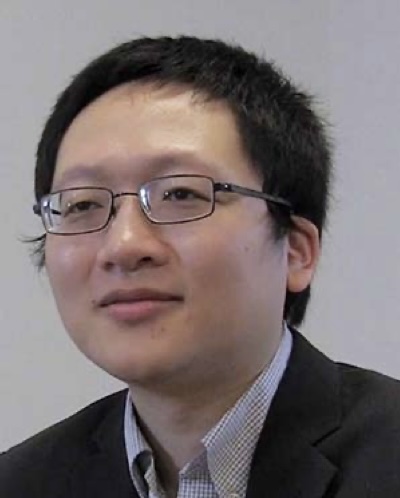}}]{Minjie Chen} (Senior Member, IEEE) received the S.M., E.E., and Ph.D. degrees in electrical engineering and computer science from Massachusetts Institute of Technology, Cambridge, MA, USA, in 2015 and the B.S. degree in electrical engineering from Tsinghua University, Beijing, China, in 2009. He is an Assistant Professor of electrical and computer engineering and the Andlinger Center for Energy and the Environment at Princeton University. His research interests include high-frequency power electronics, power architecture, power magnetics, advanced packaging, data-driven methods, design automation, and design methods of high-performance power electronics for emerging and important applications. Dr. Chen is a recipient of the IEEE PELS Richard M. Bass Outstanding Young Power Electronics Engineer Award, the Princeton SEAS E. Lawrence Keyes, Jr./Emerson Electric Co. Junior Faculty Award, the NSF CAREER Award, six IEEE Transactions on Power Electronics Prize Paper Awards, a COMPEL Best Paper Award, an ICRA Best Poster Award, an IROS Best Paper Finalist, three ECCE Best Demo Awards, a 3D-PEIM Rao R. Tummala Best Paper Award, an OCP Best Paper Award, a Siebel Research Award, a C3.ai Research Award, a First Place Award of Princeton Keller Center Innovation Forum, and the MIT EECS D. N. Chorafas Ph.D. Thesis Award. He was listed on the Princeton Engineering Commendation List for Outstanding Teaching multiple times. He is the Vice Chair of PELS TC10 Design Methodologies, and the TPC member of a few flagship PELS conferences including APEC, ECCE, COMPEL, and ICDCM.\end{IEEEbiography}
\vskip -2\baselineskip plus -1fil
\begin{IEEEbiography}[{\includegraphics[width=1in,height=1.25in,clip,keepaspectratio]{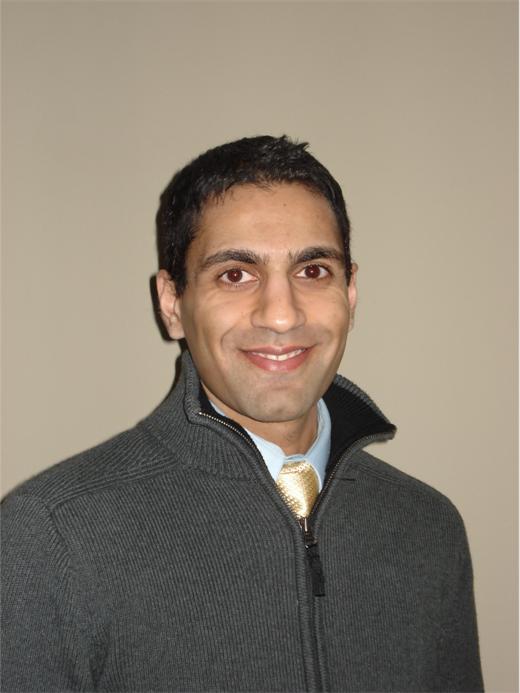}}]{Naveen Verma} (Senior Member, IEEE) received the B.A.Sc. degree in Electrical and Computer Engineering from the UBC, Vancouver, Canada in 2003, and the M.S. and Ph.D. degrees in Electrical Engineering from MIT in 2005 and 2009 respectively. Since July 2009 he has been at Princeton University, where he is currently director of the Keller Center for Innovation in Engineering Education and Professor of Electrical and Computer Engineering. 

His research focuses on advanced sensing systems, exploring how systems for learning, inference, and action planning can be enhanced by algorithms that exploit new sensing and computing technologies. This includes research on large-area, flexible sensors, energy-efficient statistical-computing architectures and circuits, and machine-learning and statistical-signal-processing algorithms.

Prof. Verma has served as a Distinguished Lecturer of the IEEE Solid-State Circuits Society, and currently serves on the technical program committees for ISSCC, VLSI Symp., DATE, and IEEE Signal-Processing Society (DISPS). Prof. Verma is recipient or co-recipient of the 2006 DAC/ISSCC Student Design Contest Award, 2008 ISSCC Jack Kilby Paper Award, 2012 Alfred Rheinstein Junior Faculty Award, 2013 NSF CAREER Award, 2013 Intel Early Career Award, 2013 Walter C. Johnson Prize for Teaching Excellence, 2013 VLSI Symp. Best Student Paper Award, 2014 AFOSR Young Investigator Award, 2015 Princeton Engineering Council Excellence in Teaching Award, and 2015 IEEE Trans. CPMT Best Paper Award. 
\end{IEEEbiography}
\vskip -2\baselineskip plus -1fil
\begin{IEEEbiography}[{\includegraphics[width=1in,height=1.25in,clip,keepaspectratio]{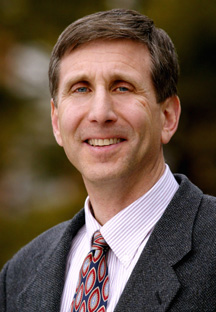}}]{James C. Sturm}(Fellow, IEEE) was born in Berkeley Heights, NJ, in 1957. He received the B.S.E. degree in electrical engineering and engineering physics from Princeton University, Princeton, NJ, in 1979 and the M.S.E.E. and Ph.D. degrees in 1981 and 1985, respectively, both from Stanford University, Stanford, CA. 

In 1979, he joined Intel Corporation, Santa Clara, CA, as a Microprocessor Design Engineer, and in 1981 he was a Visiting Engineer at Siemens, Munich, Germany. In 1986, he was joined the faculty of Princeton University, where he is currently the Stephen R. Forrest Professor of Electrical and Computer Engineering and the Chair of the Department of Electrical and Computer Engineering. From 2003 to 2015 he was the founding director of the Princeton Institute for the Science and Technology of Materials (PRISM), and from 1994 to 1995, he was a von Humboldt Fellow at the Institut fuer Halbleitertechnik at the University of Stuttgart, Stuttgart, Germany. He has worked in the fields of silicon-based heterojunctions, thin-film and flexible electronics, photovoltaics, the nano-bio interface, SOI, and three-dimensional (3–D) integration.  

He was a National Science Foundation Presidential Young Investigator. He has won over ten awards for teaching excellence.  In 1996 and 1997, he was the technical program chair and general chair of the Device Research Conference, respectively. He served on the organizing committee of IEDM (1988 to 1992 and 1998 to 1999), having chaired both the solid-state device and detectors/sensors/displays committees. He has served on the boards of directors of the Materials Research Society and the Device Research Conference, and was a co-founder of Aegis Lightwave.  
\end{IEEEbiography}

\end{document}